\documentclass[sigconf]{acmart}

\usepackage{enumitem}
\usepackage{xcolor}
\usepackage{tikz}
\usepackage{pgfplots}
\usepackage[multiple]{footmisc}
\usepackage{multirow}
\usepackage{bm}
\usepackage{booktabs}
\usepackage{hyperref}
\usepackage{array}
\usepackage{subcaption}
\usepackage{graphicx}
\usepackage{tabularx}
\usepackage{geometry}
\usepackage{lipsum}
\usepackage{algorithm}
\usepackage{algorithmicx}
\usepackage{algpseudocode}
\usepackage{soul}
\usepackage{float}
\usepackage{tcolorbox}

\newcommand{\spec}{{\it spec.}}

\newcommand{\ie}{{\textit i.e.}}
\newcommand{\eg}{{\textit e.g.}}
\newcommand{\ours}{\textsc{\textsf{CAPER}}}

\newtheorem{prob}{\textbf{Problem}}

\newcommand{\utempemb}{$\mathbf{u}_x(t_m)$}
\newcommand{\ctempemb}{$\mathbf{c}_z(t_m)$}
\newcommand{\uevolemb}{$\tilde{\mathbf{u}}_x(t_m)$}
\newcommand{\cevolemb}{$\tilde{\mathbf{c}}_z(t_m)$}
\newcommand{\pemb}{$\mathbf{p}_y$}

\newcommand{\rom}[1]{\lowercase\expandafter{\romannumeral #1\relax}}

\pgfplotsset{% https://tex.stackexchange.com/a/75811/194703
    name nodes near coords/.style={
        every node near coord/.append style={
            name=#1-\coordindex,
            alias=#1-last,
        },
    },
    name nodes near coords/.default=coordnode
}

\pgfplotsset{compat=1.5.1}
\def\addlegendimage{\csname pgfplots@addlegendimage\endcsname}
\usetikzlibrary{patterns}
\usetikzlibrary{pgfplots.groupplots}
\usetikzlibrary{
  pgfplots.colorbrewer,
}

\definecolor{aa}{rgb}{0.2,0.7,0.310}
\definecolor{cc}{rgb}{1.0,0.49,0.0}
\definecolor{bb}{rgb}{0.514,0.325,0.831}

\AtBeginDocument{%
  }

\copyrightyear{2025}
\acmYear{2025}
\setcopyright{rightsretained}
\acmConference[KDD '25]{Proceedings of the 31st ACM SIGKDD Conference on
Knowledge Discovery and Data Mining V.1}{August 3--7, 2025}{Toronto, ON,
Canada}
\acmBooktitle{Proceedings of the 31st ACM SIGKDD Conference on Knowledge
Discovery and Data Mining V.1 (KDD '25), August 3--7, 2025, Toronto, ON,
Canada}
\acmDOI{10.1145/3690624.3709329}
\acmISBN{979-8-4007-1245-6/25/08}

\begin{document}

\title{CAPER: Enhancing Career Trajectory Prediction using Temporal Knowledge Graph and Ternary Relationship}

\author{Yeon-Chang Lee}
\affiliation{
	\institution{Ulsan National Institute of Science and Technology (UNIST)}
	\city{Ulsan}
  	\country{Korea}
}
\email{yeonchang@unist.ac.kr}

\author{JaeHyun Lee}
\affiliation{
	\institution{Hanyang University}
	\city{Seoul}
  	\country{Korea}
}
\email{dldja4@hanyang.ac.kr}

\author{Michiharu Yamashita}
\affiliation{
	\institution{The Pennsylvania State University}
	\city{University Park, PA}
  	\country{USA}
}
\email{michiharu@psu.edu}

\author{Dongwon Lee}
\affiliation{
	\institution{The Pennsylvania State University}
	\city{University Park, PA}
  	\country{USA}
}
\email{dongwon@psu.edu}

\author{Sang-Wook Kim}
\authornote{Corresponding author.}
\affiliation{
	\institution{Hanyang University}
	\city{Seoul}
  	\country{Korea}
}
\email{wook@hanyang.ac.kr}

\renewcommand{\shortauthors}{Yeon-Chang Lee, JaeHyun Lee, Michiharu Yamashita, Dongwon Lee, \&  Sang-Wook Kim}

\begin{abstract}
The problem of \emph{career trajectory prediction} (CTP) aims to predict one's future employer or job position. 
While several CTP methods have been developed for this problem, we posit that none of these methods (1) jointly considers the mutual ternary dependency between three key units (\ie, user, position, and company) of a career and (2) captures the characteristic shifts of key units in career over time, leading to an inaccurate understanding of the job movement patterns in the labor market.
To address the above challenges, we propose a novel solution, named as \ours, that solves the challenges via sophisticated temporal knowledge graph (TKG) modeling.
It enables the utilization of a graph-structured knowledge base with rich expressiveness, effectively preserving the changes in job movement patterns.
Furthermore, we devise an extrapolated career reasoning task on TKG for a realistic evaluation.
The experiments on a real-world career trajectory dataset demonstrate that \ours\ consistently and significantly outperforms four baselines, two recent TKG reasoning methods, and five state-of-the-art CTP methods
in predicting one's future companies and positions--\ie, on average,  
yielding 6.80\% and 34.58\% more accurate predictions, respectively.
The codebase of \ours\ is available at \href{https://github.com/Bigdasgit/CAPER}{https://github.com/Bigdasgit/CAPER}.
\end{abstract}




\maketitle

\begin{figure}[t]
\centering
  \includegraphics[width=0.95\linewidth]{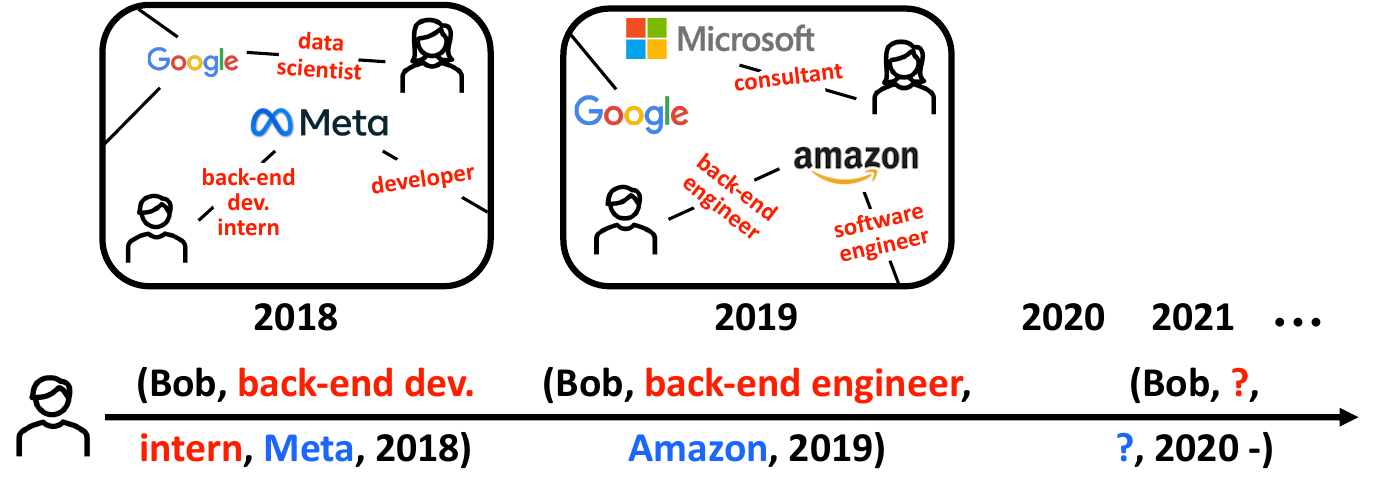} 
  \vspace{-0.2cm}
\caption{An extrapolated career reasoning task on the input temporal knowledge graph consisting of career trajectories.} \label{fig:intro}
\vspace{-0.5cm}
\end{figure}

\section{Introduction}\label{sec:introduction}

\vspace{1mm}
\noindent{\bf Background.} 
With the advent of online professional network platforms (\eg, LinkedIn and Indeed), there has been a surge of research efforts in understanding and improving diverse tasks in the labor market~\cite{BrandaoM17,OentaryoLAPOL18,chen2018linkedin,zhang2019job2vec,zhang2020large,LeeHK16,LeeLHK17}, such as 
talent recruitment \cite{QinZXZJCX18,qin2020enhanced, abdollahnejad2021deep}, talent demand/supply prediction \cite{guo2022talent,ZhuZXDX16,ZhangZSLZX21}, and career trajectory prediction \cite{li2017nemo, meng2019hierarchical, zhang2021attentive, wang2021variable}. With the fast-evolving job market and its immediate benefits to both employees and employers, in particular, the problem of {\bf Career Trajectory Prediction} ({\bf CTP}) (\ie, predict one's future employer or job position) has increasingly attracted attention in recent years \cite{li2017nemo, meng2019hierarchical, zhang2021attentive, wang2021variable}, especially when the life-long career movement has become a new norm (\eg, according to the U.S. Bureau of Labor Statistics, people in the U.S. hold an average of $8.6$ jobs from ages $18$ through $34$\footnote{\href{https://www.bls.gov/opub/ted/2022/people-born-in-early-1980s-held-an-average-of-8-6-jobs-from-ages-18-through-34.htm}{https://www.bls.gov/opub/ted/2022/people-born-in-early-1980s-held-an-average-of-8-6-jobs-from-ages-18-through-34.htm}}).
Effective solutions to the CTP problem can bring significant benefits to various groups, including governments, companies, and job seekers. 
Governments can devise better policies through an in-depth understanding of the labor market, while companies can refine their recruitment and retention strategies. 
Also, job seekers can design their own career paths more effectively by identifying promising future career opportunities. 

Motivated by such wide applicability, several CTP methods have been developed recently \cite{li2017nemo, meng2019hierarchical, zhang2021attentive, wang2021variable}. 
In a nutshell, they were designed based on the assumption that users tend to have common patterns in career movements. 
Specifically, they divide each user's career trajectory into his/her transition sequences of companies and positions. 
Then, they capture the frequent transition patterns over users with respect to companies and positions via graph recurrent neural networks (GRNN) models~\cite{GRNNsurvey2018} (\eg, long short-term memory (LSTM), gated recurrent unit (GRU), and Transformer~\cite{VaswaniSPUJGKP17}).
Finally, these methods predict a user's next career movement, including a future company and a position, based on the learned GRNN model.

\vspace{1mm}
\noindent{\bf Challenges.}
However, in this paper, we explore two challenges that have been overlooked in existing studies: 
\textbf{(C1: Career Modeling)} a mutual ternary dependency exists among \emph{three} key units of a ``career"--\ie, (\emph{user}, \emph{position}, \emph{company}, timestamp);
\textbf{(C2: Temporal Dynamics)} the characteristics per unit may change over time.

Regarding (C1), note that each career represents an intense interaction between a user, a position, and a company, which are highly related with each other.
For instance, a current career (Bob, back-end engineer, Amazon, 2023) can be regarded as an output of jointly considering Bob's past experience (\eg, computer science), Amazon's industrial characteristics (\eg, technology company), and the back-end engineer's required role (\eg, distributed programming).
Meanwhile, only one method~\cite{zhang2021attentive} attempted to directly model mutual dependency between \textit{two units}, \ie, company and position. 
However, omitting users in modeling careers is not a good idea since users often play a key role in career decisions~\cite{gati2019decision}.

Regarding (C2), it is important to consider the \textit{timestamp}, in which each career occurred, to address the dynamic nature of the labor market. 
Since each user's experience evolves over time, his/her different careers at different timestamps should be regarded as the consequence of the change in his/her characteristic.
Suppose that Bob started his career at Meta as a back-end developer intern in 2018 and then moved to Amazon as a back-end engineer in 2019. 
This suggests that Bob's internship experience at Meta influenced (partially) his move to Amazon.
Furthermore, each company's characteristics can also alter over time due to various societal/economic issues (\eg, changes in industry trends).
For instance, Meta started as a social-media-service business in 2004, but since 2021, its main focus has (reportedly) shifted to the metaverse. 
However, existing methods~\cite{li2017nemo, meng2019hierarchical, zhang2021attentive, wang2021variable} use the temporal dimension to only identify each user's career sequence, ignoring the characteristic shifts of key units in career over time.

\vspace{1mm}
\noindent{\bf Proposed Ideas.}
To address the above challenges, 
we propose a novel solution, named as \textbf{\ours} ({\bf CA}reer trajectory {\bf P}rediction approach based on t{\bf E}mporal knowledge g{\bf R}aph), which jointly learns the mutual dependency for the ternary relationships and the characteristic shifts over time. 
It consists of three key modules:

\vspace{-0.1cm}
\begin{itemize}[leftmargin=*]
    \item {\textbf{Modeling Career Trajectories (for (C1) and (C2))}}: We model the career trajectories as a form of a knowledge graph (KG), preserving the ternary relationships on career information.
    To add the temporal dimension, we span the KG as a sequence of temporal KG ({\bf TKG}) snapshots with timestamps, each snapshot of which contains the careers that co-occur at a timestamp.
    \item {\textbf{Learning Mutual Dependency (for (C1))}}: Given each KG snapshot, we obtain users, positions, and companies as low-dimensional vectors (\ie, embeddings) by aggregating their career information in the input KG snapshot.
    Here, we represent the same user and company at different timestamps as different temporal embeddings, while representing the position as a unified embedding regardless of timestamps. We assume that, unlike the user and company units, the position unit has static properties, which will be discussed in Sections~\ref{sec:approach} and~\ref{sec:evaluation}.
    \item {\textbf{Learning Temporal Dynamics (for (C2))}}: Based on temporal embeddings, we capture the changes in the characteristics of users and companies between previous and current timestamps. 
\end{itemize}

Finally, we learn the embeddings of users, companies, and positions so that the likelihood of each career (\ie, ground truth) at each timestamp is maximized.
We show experimentally that (1) our modules successfully solve the challenges of existing methods and (2) \ours\ substantially outperforms the state-of-the-art CTP methods, including NEMO~\cite{li2017nemo}, HCPNN~\cite{meng2019hierarchical}, AHEAD~\cite{zhang2021attentive}, TACTP~\cite{wang2021variable}, and NAOMI~\cite{yamashita2022looking}. 
For a realistic evaluation, we devise an additional extrapolation reasoning task (see Figure~\ref{fig:intro}) that aims to forecast the next careers that are likely to occur at the non-existent future timestamps in the input TKG, which will be elaborated in Section~\ref{sec:approach}.

\vspace{1mm}
\noindent{\bf Contributions.} 
Our contributions are as follows:
\begin{itemize}[leftmargin=*]
    \item {\textbf{New Perspective}}: We model the career trajectories of users as a TKG and then formulate the CTP problem as the \textbf{extrapolated reasoning task on TKG}. 
    We are the first who leverages the rich expressiveness of a graph-structured knowledge base for ternary relationships to address the CTP problem. 
    \item {\textbf{Effective CTP Approach}}: \ours\ jointly learns the mutual dependency for the ternary relationships and the characteristic shifts over time via \textbf{TKG modeling}, thereby successfully addressing the challenges of existing methods.
    \item {\textbf{Comprehensive Validation}}: Through extensive experiments using a real-world career trajectory dataset, we show that \ours\ consistently and significantly outperforms the state-of-the-art CTP methods, yielding \textbf{6.80\% and 34.58\% more-accurate company and position prediction on average}, 
    respectively, compared to the best competitors.
\end{itemize}

\section{Related Work}\label{sec:related_work}

\vspace{1mm}
\noindent{\bf CTP Methods.} 
Career trajectory prediction (CTP) is an essential task of predicting one’s next position and company~\cite{tarique2010global,Qin23survey}. 
\citet{li2017nemo} proposed a way to encode different types of entities (\eg, employees, skills, and companies) by integrating the profile context as well as career path dynamics. 
\citet{meng2019hierarchical} leveraged a hierarchical neural network with the attention mechanism to jointly characterize internal and external career transition.
\citet{wang2021variable} devised a unified time-aware framework that predicts not only the next company and position but also one's transition timing.
\citet{zhang2021attentive} utilized the mutual dependency for pairwise relationships of position and company via heterogeneous graph modeling.
Lastly, \cite{yamashita2022looking} predicts a future career pathway via the Transformer mechanism~\cite{VaswaniSPUJGKP17} and neural collaborative reasoning by jointly utilizing graph and BERT~\cite{ReimersG19} embeddings.
In contrast to these previous works, our work proposes a TKG-enhanced CTP approach that can jointly learn the mutual dependency for ternary relationships and the characteristic shifts over time.
The main differences between the existing CTP methods and \ours\ will be elaborated in Section~\ref{sec:discussion}.

Meanwhile, several attempts~\cite{QinZXZJCX18,YaoZQWZX22,Bian0ZZHSZW20,BianZSZW19,Fang0ZYZZZX23} have been made to model the matching between a job posting and a resume, which can be applied to the job or talent recommendation problems. 
However, as discussed in~\cite{Qin23survey}, we believe that these problems have different goals from that of the CTP problem, which requires an understanding of the underlying job movement patterns in the labor market.

\vspace{1mm}
\noindent{\bf TKG Embedding Methods.}
Temporal knowledge graph embedding (TKGE) methods aim to represent entities and relations in a given TKG into low-dimensional embeddings, which preserve the structural properties in the TKG. 
According to \cite{SunZMH021, park2022evokg}, recent TKGE methods were designed for the purpose of addressing an interpolation task \cite{JungJK21, THOR,XuCNL21,MontellaRH21,abs-2112-07791,ZhangLLFZW23, LeeLLK24} or an extrapolation task~\cite{zhu2021learning, HanCMT21,LiJLGGSWC21, LiJGLGWC20, SunZMH021, park2022evokg,ChenXS0D23}. 
Our extrapolated career reasoning task is inspired by the extrapolation task in the TKGE research.
However, they are inappropriate for the career trajectory dataset, failing to fully leverage the intrinsic characteristics of the job market.

\begin{figure*}[t]
\centering
  \includegraphics[width=.97\linewidth]{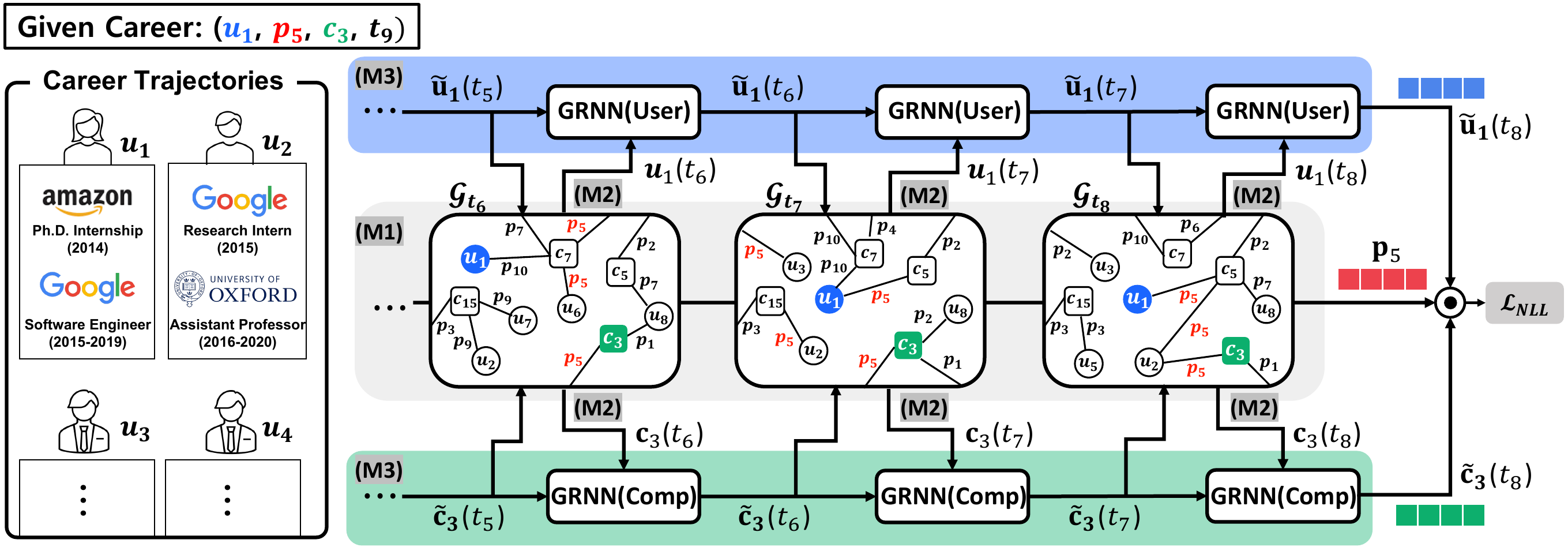} 
\vspace{-0.1cm}
\caption{Overview of \ours, which consists of three key modules: (M1) modeling career trajectories, (M2) learning mutual dependency, and (M3) learning temporal dynamics. Given a career \boldsymbol{$(u_1,p_5,c_3,t_9)$}, \ours\ infers the corresponding career's likelihood using the temporal and evolution embeddings of \boldsymbol{$u_1$} and \boldsymbol{$c_3$} before \boldsymbol{$t_9$} and the embedding of \boldsymbol{$p_5$}.}
\vspace{-0.1cm}
\label{fig:overview}
\end{figure*}

\section{Preliminaries}\label{sec:preliminaries}

\subsection{Data Description}\label{sec:pre_data}

In this study, we obtained a real-world career trajectory dataset from a global career platform, FutureFit AI\footnote{https://www.futurefit.ai}.
From this dataset, we randomly extracted resumes (\ie, users) with at least {\em five} valid career transitions in the U.S. from 1968 to 2020. In total, we obtained 333K+ valid resumes. 
Each resume includes a user-related information (\eg, user ID, demographics, and education history),  career-transition information (\eg, companies and positions), and timestamps (\ie, starting and ending date of each work experience). 
Any private elements in resumes were pre-anonymized. 
Table \ref{table:dataset} summarizes statistics about our final dataset after pre-processing.
The pre-processing procedures, additional statistics, and a sharing statement are presented in the Appendix.

\begin{table}[t]
\centering
\footnotesize
\caption{Statistics of the career trajectory dataset 
}
\vspace{-0.25cm}
\label{table:dataset}
\renewcommand{\arraystretch}{1.1}
\resizebox{.45\textwidth}{!}{
\begin{tabular}{c|rrr}
\toprule
 & \multicolumn{1}{r}{\textbf{All}} & \multicolumn{1}{r}{\textbf{Training}} & \multicolumn{1}{r}{\textbf{Test}} \\ \midrule
\textbf{$|$Users$|$} &  333,399 & 333,399  & 183,575 \\ \midrule
\textbf{$|$Companies$|$} & 9,000 & 9,000 & 8,510  \\ 
\textbf{$|$Positions$|$} & 425 & 425 & 421  \\
\midrule
\textbf{$|$Careers$|$} & 8,218,162 & 6,453,546 & 1,764,616  \\
\textbf{Avg. \# Careers per User}  & 24.64 & 19.35 & 9.61  \\  \midrule
\textbf{Period} & 1968 $\sim$ 2020 & 1968 $\sim$ 2015  & 2016 $\sim$ 2020  \\
\textbf{Granularity} & year & year  & year \\
\bottomrule
\end{tabular}
}
\vspace{-0.3cm}
\end{table}

\subsection{The Problem Formulation: CTP}\label{sec:pre_problem}
% We formulate the problem of career trajectory prediction. 
Let $\mathcal{U}=\{u_1,u_2,\cdots,u_i\}$, $\mathcal{P}=\{p_1,p_2,\cdots,p_j\}$, $\mathcal{C}=\{c_1,c_2,\cdots,c_k\}$, and $T=\{t_1,t_2,\cdots,t_l\}$ denote sets of $i$ users, $j$ positions, $k$ companies, and $l$ timestamps, respectively. 
In our dataset, the granularity of a timestamp is the year.
Also, we denote a set of career trajectories across all users as $\mathcal{D}=\{\mathcal{D}_{u_1},\mathcal{D}_{u_2},\cdots,\mathcal{D}_{u_i}\}$.
Each user $u_x$'s 
trajectory is denoted as $\mathcal{D}_{u_x}=\{(u_x,p_{(x,1)},c_{(x,1)},$ $t_{(x,1)}), (u_x,p_{(x,2)},c_{(x,2)},t_{(x,2)}),\cdots$, $ (u_x,p_{(x,s_x)},c_{(x,s_x)},t_{(x,s_x)})\}$, \\where $(u_x,p_{(x,n)},c_{(x,n)},t_{(x,n)})$ represents the $n$-th career of $u_x$, \ie, $u_x$ worked in a company $c_{(x,n)}$ with a position $p_{(x,n)}$ at a timestamp $t_{(x,n)}$, and $s_x$ indicates the total length of the working interval for $u_x$ (\ie, the difference between the first and the last timestamps in $\mathcal{D}_{u_x}$).
Then, our problem can be defined as follows~\cite{yamashita2022looking}: 
\begin{prob}[\textbf{Career Trajectory Prediction (CTP)}]\label{prob:ctp}
Given a query user $u_x$ and her/his career trajectory $\mathcal{D}_{u_x}$, predict the next $M$ careers of $u_x$, including $M$ companies $\{c_{(x,{s_x}+1)},c_{(x,{s_x}+2)},\cdots,c_{(x,{s_x}+M)}\}$ and $M$ positions $\{p_{(x,{s_x}+1)},p_{(x,{s_x}+2)},\cdots,p_{(x,{s_x}+M)}\}$.
\end{prob}

Next, we formulate the CTP problem as an extrapolated career reasoning task on a {\em temporal knowledge graph} ({\bf TKG}).
Let $\mathcal{G}=\{\mathcal{G}_{t_1},\mathcal{G}_{t_2},\cdots,\mathcal{G}_{t_l}\}$ denotes a TKG~\cite{shaoxiong2022kgsurvey, rakshit2017knowevolve}
that consists of a sequence of knowledge graph (KG) snapshots with timestamps based on career trajectories. 
Each KG snapshot $\mathcal{G}_{t_m}=\{(u_x,p_y,c_z,t_m)\}$ at $t_m$ is a bidirected multi-relational graph and stores the careers that co-occur at $t_m$.
For a career $(u_x,p_y,c_z,t_m)$, in a KG snapshot $\mathcal{G}_{t_m}$, the user $u_x$ (\ie, subject entity) and the company $c_z$ (\ie, object entity) are represented as nodes, and the position $p_y$ (\ie, relation) is represented as a label of an (undirected) edge between $u_x$ and $c_z$ (see Figure~\ref{fig:intro}).
Such a TKG modeling to career trajectories will be more elaborated in Section~\ref{sec:sec-modules}. 
Then, the CTP problem can be restated to the {\em extrapolated career reasoning} task as follows:
\begin{prob}[\textbf{Extrapolated Career Reasoning}]\label{prob:ert}
Given a TKG $\mathcal{G}=\{\mathcal{G}_{t_1},\mathcal{G}_{t_2},\cdots,\mathcal{G}_{t_l}\}$ and a query user $u_x$, predict the unknown (next) careers 
of $u_x$ that are likely to occur at future $M$ timestamps $\{t_{(l+1)}, t_{(l+2)}, \cdots, t_{(l+M)}\}$ in the input TKG $\mathcal{G}$.
\end{prob}

\begin{table}[t]
  \centering
  \small
  \caption{Notations used in this paper}
  \label{Notations}
  \vspace{-0.2cm}
\resizebox{.48\textwidth}{!}{
  \renewcommand{\arraystretch}{1.3}
    \begin{tabular}{p{0.3\linewidth}|p{0.75\linewidth}} \toprule
    \hfil \textbf{Notation}                                                           & \hfil \textbf{Description} \\ \midrule    
    \hfil $\mathcal U$, $\mathcal P$, $\mathcal C$, $\mathcal T$                                                       &  Sets of users, positions, companies, and timestamps \\ % \midrule
    \hfil $u_x,p_y,c_z,t_m$ & User (\ie, subject entity), position (\ie, relation), company (\ie, object entity), and timestamp \\
    \hfil $\mathcal{D}$, $\mathcal{D}_{u_x}$ & Set of career trajectories and $u_x$'s career trajectory \\
    \hfil $p_{(x,n)},c_{(x,n)}$ & $n$-th position and company of $u_x$ in $\mathcal{D}_{u_x}$ \\ \midrule
    \hfil $\mathcal G, \mathcal G_{t_m}$                                                                         &  TKG and KG snapshot w.r.t $t_m$ \\ %\midrule
    \hfil  \utempemb, \ctempemb  & Temporal embeddings of $u_x$ and $c_z$ w.r.t $t_m$ \\ %\midrule
    \hfil  \uevolemb, \cevolemb  & Evolution embeddings of $u_x$ and $c_z$ w.r.t $t_m$ \\ %\midrule
    \hfil  \pemb  & Embedding of $p_y$ \\ \midrule 
    \hfil $L$ & Number of GCN Layers  \\
    \hfil $\mathcal N_{u_x}(\mathcal G_{t_m})$ & Set of pairs of a company $c_{z^\prime}$ and a position $p_{y^\prime}$
    % \michi{is this $p_{y^\prime}$?} 
    included in the careers, which include $u_x$, at $t_m$ \\
    % in $\mathcal G$ \\
    \hfil $\mathcal N_{c_z}(\mathcal G_{t_m})$ & Set of pairs of a user $u_{x^\prime}$ and a position
    $p_{y^\prime}$ included in the careers, which include $c_z$, at $t_m$ \\ 
    \bottomrule
  \end{tabular}
  }
  \vspace{-0.4cm}
  \label{table:notations}
\end{table}

Table~\ref{table:notations} summarizes a list of notations used in this paper.

\section{The Proposed Approach: \ours}\label{sec:approach}

\subsection{Overview}\label{sec:sec-overview}

In this section, we present the overall procedure of \ours. 
As shown in Figure~\ref{fig:overview}, \ours\ performs the following three key modules: (M1) modeling career trajectories, (M2) learning mutual dependency, and (M3) learning temporal dynamics.

Given a set $\mathcal{D}$ of career trajectories across all users, in (M1), \ours\ models them as a sequence of KG snapshots with timestamps, $\mathcal{G}=\{\mathcal{G}_{t_1},\mathcal{G}_{t_2},\cdots,\mathcal{G}_{t_l}\}$.
Then, \ours\ represents the entities (\ie, users and companies) and relations (\ie, positions) in $\mathcal{G}$ as $d$-dimensional embeddings.
Specifically, it constructs the same user $u_x$ and company $c_z$ at different timestamps $t_m$ as different temporal embeddings \utempemb\ and \ctempemb, respectively, while constructing the same position $p_y$ as a single unified embedding \pemb\ regardless of timestamps. 
The intuition behind this design choice is to differentiate the dynamic natures of users and companies from the static nature of positions, which will be elaborated in Section~\ref{sec:sec-modules}.

Next, \ours\ repeatedly performs (M2) and (M3) whenever each KG snapshot in $\mathcal{G}$ is given sequentially. 
Given a KG snapshot $\mathcal{G}_{t_m}$, in (M2), \ours\ updates the temporal embedding \utempemb\ of $u_x$ with respect to $t_m$ so that the embedding can preserve the mutual dependencies on his/her career; in the same manner, it updates \ctempemb\ of $c_z$ with respect to $t_m$.
Using \utempemb, in (M3), \ours\ obtains the evolution embedding \uevolemb\ of $u_x$ with respect to $t_m$, which captures the characteristic shift of $u_x$ between the previous  and current timestamps $t_{m-1}$ and $t_{m}$; in the same manner, it obtains \cevolemb\ of $c_z$ with respect to $t_m$ using \ctempemb.
Then, \uevolemb\ and \cevolemb\ with respect to $t_m$ are used to initialize the temporal embeddings $\mathbf{u}_x(t_{m+1})$ and  $\mathbf{c}_z(t_{m+1})$ with respect to $t_{(m+1)}$.

Lastly, given a query career $(u_x,p_y,c_z,t_{m+1})$ in $\mathcal{G}_{t_{m+1}}$, \ours\ learns the temporal and evolution embeddings of $u_x$ and $c_z$ before $t_{m+1}$ and the embedding of $p_y$ via the negative log-likelihood (NLL) loss so that the corresponding career's likelihood can be maximized.

\begin{figure}[t]
\centering
  \includegraphics[width=0.93\linewidth]{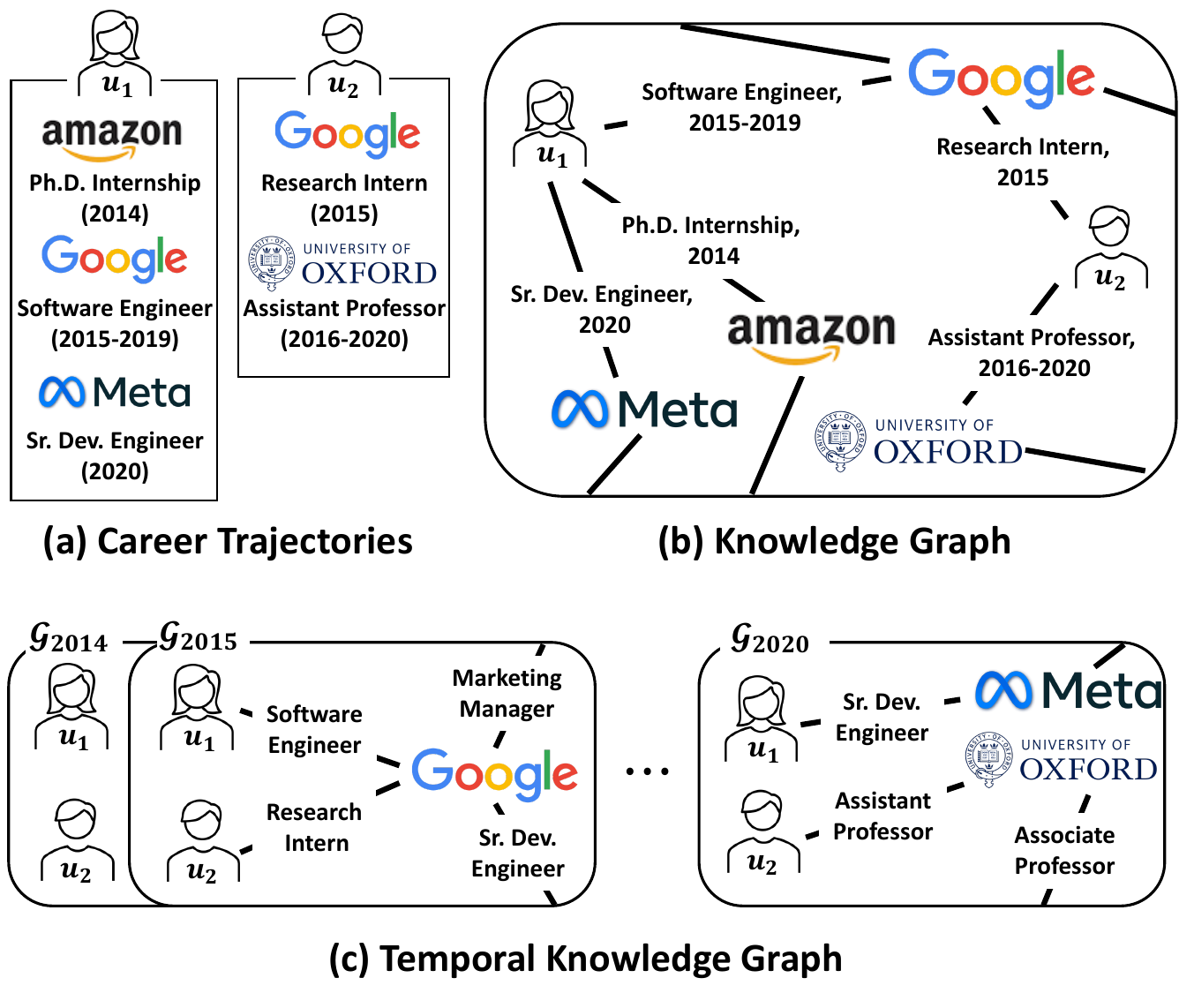} 
  \vspace{-0.2cm}
\caption{A toy example of modeling career trajectories.
}\label{fig:modeling}
\vspace{-0.2cm}
\end{figure}

\subsection{Key Modules}\label{sec:sec-modules}

\noindent{\bf (M1) Modeling Career Trajectories.}
This module aims to model the given career trajectories as a graph structure that can simultaneously represent the mutual dependency for the ternary relationships on a user, a position, and a company (\ie, (C1)) and the changes in characteristics of users and companies over time (\ie, (C2)).

To address (C1), we first construct the career trajectories as a KG, where the users $u_x$ and companies $c_z$ are represented as nodes, and the positions $p_y$ are represented as labels of the edges between $u_x$ and $c_z$. 
Suppose that there are the following career trajectories of two users $u_1$ and $u_2$ (see Figure~\ref{fig:modeling}-(a)): $\mathcal{D}_{u_1}=\{(u_1,$ Ph.D. internship, Amazon, $2014),$ 
$(u_1,$ software engineer, Google, $2015), \cdots,(u_1,$ software engineer, Google, $2019),
(u_1,$ senior develop engineer, Meta, $2020)\}$ and $\mathcal{D}_{u_2}=\{(u_2,$ research intern, Google, $2015),$ 
$(u_2,$ assistant professor, University of Oxford, $2016), \cdots,(u_2,$ assistant professor, University of Oxford, $2020)\}$. 
% In this case, 
Figure~\ref{fig:modeling}-(b) shows a KG constructed based on the career trajectories. 
By doing so, we can successfully preserve the intense interaction on three highly-correlated units, \ie, a user, a company, and a position, which form each career.
The built KG only contains the actual careers of the users.

To address (C2), we then span the input KG as a sequence of KG snapshots with timestamps, each of which contains the careers that co-occur at a specific timestamp. 
Figure~\ref{fig:modeling}-(c) shows KG snapshots with timestamps constructed by the aforementioned career trajectories of two users.
For example, $\mathcal{G}_{2015}$ includes both careers $(u_1,$ software engineer, Google, $2015)$ and $(u_2,$ research intern, Google, $2015)$ that co-occur at $2015$.
By doing so, we can separately preserve the mutual dependency on careers that co-occur at a specific timestamp, capturing characteristic shifts over time.
To the best of our knowledge, none of the existing methods consider the fact that the characteristics of some units can alter over time.

\vspace{1mm}
\noindent{\bf (M2) Learning Mutual Dependency.}
This module aims to reflect the mutual dependency on careers that co-occur at $t_m$ into temporal embeddings of users and companies with respect to $t_m$.
As mentioned above, the intuition behind this design choice is that each career reveals a high correlation between a user, a position, and a company in the corresponding career.

Therefore, we design a timestamp-aware graph convolutional networks (GCN) that learns the mutual dependency on careers at a timestamp.
\ours\ obtains each user $u_x$'s (resp. each company $c_z$'s) temporal embedding \utempemb\ (resp. \ctempemb) with respect to $t_m$ by aggregating career information of $u_x$ (resp. $c_z$) at $t_m$:
\begin{equation}\label{eq:tgcn-user}
\begin{aligned}
\mathbf{u}_x(t_m)^{l} &= \mathbf{u}_x(t_m)^{l-1} +\sum_{(c_{z^\prime},p_{y^\prime})\in{\mathcal N_{u_x}(\mathcal G_{t_m})}} \frac{1}{c}(\mathbf{c}_{z^\prime}(t_{m})^{l-1} \circ \textbf{p}_{y^\prime}),
\end{aligned}
\end{equation}
\begin{equation}\label{eq:tgcn-comp}
\begin{aligned}
\mathbf{c}_z(t_m)^{l} &= \mathbf{c}_z(t_m)^{l-1} +\sum_{(u_{x^\prime},p_{y^\prime})\in{\mathcal N_{c_z}(\mathcal G_{t_m})}} \frac{1}{c}(\mathbf{u}_{x^\prime}(t_{m})^{l-1} \circ \textbf{p}_{y^\prime}),
\end{aligned}
\end{equation}
where $\mathbf{u}_x(t_m)^{0}$, $\mathbf{c}_z(t_m)^{0}$, $\mathbf{u}_{x^\prime}(t_{m})^{0}$, $\mathbf{c}_{z^\prime}(t_{m})^{0}$, and $\mathbf{p}_{y^\prime}$ are randomly initialized without using any of their handcrafted features.
In addition, $\circ$ and $l\in \{1,\cdots, L\}$ represent the element-wise product and the $l$-th \texttt{GCN} layer, respectively.
Also, $\mathcal N_{u_x}(\mathcal G_{t_m})$ represents a set of company-position pairs $(c_{z^\prime},p_{y^\prime})$ included in $u_{x}$'s careers at $t_m$.
Similarly, $\mathcal N_{c_z}(\mathcal G_{t_m})$ represents a set of user-position pairs $(u_{x^\prime},p_{y^\prime})$ included in $c_{z}$'s careers at $t_m$.
In addition, $c$ is a normalization constant, which is normally defined by the degree of nodes in GCNs~\cite{shaoxiong2022kgsurvey,KongKJ0LPK22,KimLSK22,KimLK23,SharmaLNSSKK24}.
Finally, \ours\ considers $\mathbf{u}_x(t_m)^{L}$ (resp. $\mathbf{c}_z(t_m)^{L}$) obtained from the $L$-th \texttt{GCN} layer as $u_x$’s (resp. $c_z$'s) final temporal embedding \utempemb\ (resp. \ctempemb) with respect to $t_m$.

Here, we note that, unlike companies, users have few careers at each timestamp.
On our real-world dataset, we confirmed that in each KG snapshot $\mathcal{G}_{t_{m}}$, each company $c_z$ is included in 29.78 careers on average,
while each user only has 2.37 careers on average.\footnote{As shown in Figure~\ref{fig:overview}, when a user $u_1$'s total length of the working interval is 3 and he
/she has four careers for 3 years (\ie, $\mathcal{D}_{u_1}=\{$$(u_1,p_{10},c_7,t_6)$, $(u_1,p_{10},c_7,t_7)$,  $(u_1,p_5,c_5,t_7)$, $(u_1,p_5,c_5,t_8)\}$), the average number of $u_1$'s careers per timestamp becomes $1.33$ (\spec, (1 (for $t_6$) + 2 (for $t_7$) +1 (for $t_8$)) / 3).}
In this case, accordingly, there remains a question of whether each user's temporal embedding with respect to each timestamp is informative. 
Basically, we admit that each user's temporal embedding with respect to the timestamp, where he/she had the first career, may not be that informative.
However, we believe that like a user's experience evolves while increasing his/her working interval over time, it is a natural way that the user's temporal embeddings also evolve over time gradually and thus become more informative. 

Furthermore, recall that each position is modeled as a label of an edge between two entities of a user and a company in the input TKG. 
Therefore, following the existing TKG-based GCN mechanism (\eg, \cite{THOR,LiJLGGSWC21}),
we only use each position embedding as the aggregated information when obtaining the temporal embeddings of users and companies included in the corresponding position's careers.

\vspace{1mm}
\noindent{\bf (M3) Learning Temporal Dynamics.}
This module aims to capture the characteristic shifts of users and companies between the previous timestamp $t_{m-1}$ and the current timestamp $t_m$.
The intuition behind this design choice is that each user's experience evolves over time and each company's characteristics can also change over time~\cite{JinLS0SDK23,YooLSK23,KimLK24}. 
On the other hand, we assume that most positions have static properties so that their characteristics do not change significantly even over time.\footnote{One can still argue that the characteristics of some positions may change significantly over time, \eg, due to advances in technology, the role of data scientists is expanding to a wide variety of fields. 
However, it is not trivial to identify such individual positions. Thus, we leave it as future work.}
That's why we do not consider characteristic shifts of positions in (M3).
We experimentally show in Section~\ref{sec:evaluation} that not considering the characteristic shifts of positions is more effective than considering them.

We design a GRNN that learns the evolution patterns of users and companies over time.
Specifically, \ours\ obtains each user $u_x$'s (resp. each company $c_z$'s) evolution embedding \uevolemb\ (resp. \cevolemb) with respect to $t_m$ as follows: 
\begin{equation}\label{eq:grnn-user}
\begin{aligned}
\tilde{\mathbf{u}}_x(t_m) &= \texttt{GRNN} (\tilde{\mathbf{u}}_x(t_{m-1}),\mathbf{u}_x(t_m)),
\end{aligned}
\end{equation}
\begin{equation}\label{eq:grnn-comp}
\begin{aligned}
\tilde{\mathbf{c}}_z(t_m) &= \texttt{GRNN}(\tilde{\mathbf{c}}_z(t_{m-1}),\mathbf{c}_z(t_m)),
\end{aligned}
\end{equation}
where \texttt{GRNN} indicates a recurrent function;
\ours\ can employ any functions, such as recurrent neural network (RNN), gated recurrent unit (GRU), and long short-term memory (LSTM).
Also, $\tilde{\mathbf{u}}_x(t_{m-1})$ and $\tilde{\mathbf{c}}_z(t_{m-1})$ represents the evolution embedding of $u_x$ and $c_z$, respectively, with respect to the previous timestamp $t_{m-1}$.
Here, $\tilde{\mathbf{u}}_x(t_{({u_x},0)})$ with respect to $t_{({u_x},0)}$ (\ie, a timestamp right before $u_x$'s first timestamp) is randomly initialized. 
In the same manner, $\tilde{\mathbf{c}}_z(t_{({c_z},0)})$ with respect to $t_{({c_z},0)}$ is randomly initialized as well.
See the Appendix for a detailed implementation of \texttt{GRNN}.

\subsection{Training and Inference}\label{sec:sec-training}

\textbf{Training.} Given each career $(u_x,p_y,c_z,t_{m+1})$ in $\mathcal{G}_{t_{m+1}}$, \ours\ obtains evolution embedding \uevolemb\ of $u_x$ with respect to $t_m$, evolution embedding \cevolemb\ of $c_z$ with respect to $t_m$, and embedding \pemb\ of $p_y$. 
Then, the likelihood between $u_x$ and $c_z$ and that between $u_x$ and $p_y$ are defined by the following two score functions:
\begin{equation}\label{eq:score}
\begin{aligned}
\phi(u_x,c_z,t_{m+1}) &= \tilde{\mathbf{u}}_x(t_{m}) \cdot \tilde{\mathbf{c}}_z(t_{m}), \\
\phi(u_x,p_y,t_{m+1}) &= \tilde{\mathbf{u}}_x(t_{m}) \cdot \mathbf{p}_y, \\
\end{aligned}
\end{equation}
where ``$\cdot$" represents the dot product. 

\ours\ learns the embeddings of users, positions, and companies by optimizing the following negative log-likelihood (NLL) loss:
\begin{equation}\label{eq:nll}
\begin{aligned}
&\mathcal{L} = -\sum_{m=1}^{|T|}\sum_{(u_x,p_y,c_z,t_{m})\in\mathcal G_{t_{m}}}\log p(c_z|u_x,t_{m})+\log p(p_y|u_x,t_{m}),
\end{aligned}
\end{equation}
\begin{equation}\label{eq:likelihood}
\begin{aligned}
&p(c_z|u_x,t_{m}) = \frac{\text{exp}(\phi(u_x,c_z,t_{m}))}{\sum\nolimits_{c_{z^\prime}}\text{exp}(\phi(u_x,c_{z^\prime},t_{m}))},\\
&p(p_y|u_x,t_{m}) = \frac{\text{exp}(\phi(u_x,p_y,t_{m}))}{\sum\nolimits_{p_{y^\prime}}\text{exp}(\phi(u_x,p_{y^\prime},t_{m}))},\\
\end{aligned}
\end{equation}
where $p(c_z|u_x,t_{m})$ and $p(p_y|u_x,t_{m})$ represent the likelihood between $u_x$ and $c_z$ and that between $u_x$ and $p_y$ with respect to $t_{m}$, respectively. 
Also, $c_{z^\prime}$ and $p_{y^\prime}$ represent the companies and positions, respectively, which have never been included in $u_x$’s careers before $t_{m}$.
Intuitively, the goal of the NLL loss is to achieve a high likelihood for each existent career and a low likelihood for each non-existent career.
By doing so, \ours\ can accurately preserve the mutual dependency of units and their characteristic shifts over time in the embedding space. 

\vspace{1mm}
\noindent\textbf{Multi-Step Inference.} 
Given a TKG $\mathcal{G}=\{\mathcal{G}_{t_1},\mathcal{G}_{t_2},\cdots,\mathcal{G}_{t_l}\}$, a query user $u_x$, and the number of future timestamps to predict $M$, \ours\ sequentially predicts the top-$k$ companies $c_z$ and top-$k$ positions $p_y$ of $u_x$ that are most likely to occur per future timestamp, \ie, $t_{(l+1)}, t_{(l+2)}, \cdots, t_{(l+M)}$, via Eq.~(\ref{eq:score}).
To this end, we obtain the temporal and evolution embeddings of $u_x$ and all companies $c_z$ with respect to each future timestamp, and the embeddings of all positions $p_y$ by performing \texttt{GCN} and \texttt{GRNN} encoders repeatedly.

\ours\ requires no \textit{additional} training during this multi-step inference process. 
Instead, since we do not know the career trajectories after $t_l$, the only extra task is to infer KG snapshots with respect to future timestamps that can be used for the \texttt{GCN} encoder.
To obtain the temporal embeddings with respect to $t_{(l+1)}$, we use an already-built KG snapshot $\mathcal{G}_{t_l}$ with respect to $t_l$, which contains the \textit{actual careers} of the users.
In contrast, to obtain the temporal embeddings after $t_{(l+1)}$, we sequentially construct KG snapshots, \ie, $\mathcal{G}_{t_{(l+1)}}$, $\mathcal{G}_{t_{(l+2)}}$, $\cdots$, $\mathcal{G}_{t_{(l+(M-1))}}$, based on each user's \textit{inferred careers}, which consist of two pairs, \ie, (1st ranked company, 1st ranked position) and (2nd ranked company, 1st ranked position), between his/her top-2 companies and his/her top-1 position predicted at the previous timestamp. 
This design choice is based on the following observations: (1) the number of companies is much larger than the number of positions, as shown in Table~\ref{table:dataset}, and (2) each user has an average of 2.37 careers in each KG snapshot.

As in other CTP work~\cite{meng2019hierarchical,zhang2021attentive,wang2021variable, yamashita2022looking}, this paper does not assume the \textit{inductive setting} typically used to address the prediction problem for new users, companies, and positions. 
However, as \ours\ is designed based on KG-based \texttt{GCN} encoders, it can simply leverage the inductive ability~\cite{TrivediFBZ19, XuRKKA20} inherent in GCNs by utilizing embeddings of existing neighbor nodes for new nodes.
We leave the sophisticated design for an inductive setting as future work.

\subsection{Discussion}\label{sec:discussion}

\subsubsection{Novelty}
We further highlight the novelty of \ours\ in contrast to state-of-the-art CTP methods~\cite{meng2019hierarchical,zhang2021attentive,wang2021variable, yamashita2022looking}.

\vspace{1mm}
\noindent\textbf{Mutual Dependency.} 
NAOMI~\cite{yamashita2022looking} preserves the structural dependency between positions or companies, while AHEAD~\cite{zhang2021attentive} preserves the mutual dependency for pairwise relationships between each position and each company, disregarding ternary relationships in both cases.
Notably, AHEAD~\cite{zhang2021attentive} represents the career trajectories as a heterogeneous graph, where nodes indicate companies and positions, and edges indicate the career transitions between two companies or positions and whether a position belongs to a company~\cite{zhang2021attentive}. 
However, we claim that omitting users in  modeling careers is not a good idea since users often play a key role in career decisions~\cite{gati2019decision}.
Conversely, both HCPNN~\cite{meng2019hierarchical} and TACTP~\cite{wang2021variable} model a user's company sequence by leveraging job sequences and/or features (\eg, number of social connections and self-introduction), and differ significantly from \ours\ in that: (1) they fail to accurately represent the position and user embeddings, as they only model users' company sequences; (2) they are unable to benefit from the graph structure, which includes high-order relationships (\ie, more than 2 hops) among users, companies, and positions.
Thus, we argue that none of the existing methods jointly model explicit and implicit correlation among the three units of a user, a position, and a company, which is critically important.

\vspace{1mm}
\noindent\textbf{Characteristic Shifts.} 
Existing methods of~\cite{meng2019hierarchical,zhang2021attentive,wang2021variable,yamashita2022looking} employed RNN encoders to learn the career sequences of users.
This design aims to capture \textit{common patterns in career movements} among users, rather than to capture the evolution in both of users and companies over time.
Under such a design, each company is represented as a \textit{single} learnable embedding~\cite{meng2019hierarchical,wang2021variable,zhang2021attentive,yamashita2022looking}, while each user is represented as a \textit{single static} (\ie, non-learnable) feature~\cite{meng2019hierarchical,wang2021variable}.
Accordingly, they cannot capture the characteristics of each user/company at each timestamp, being unable to model their evolution.
Thus, we argue that none of the existing methods consider the characteristic shifts of users and companies over time.

\subsubsection{Complexity}

\ours\ employs \texttt{GCN} and \texttt{GRNN} to learn mutual dependency and temporal dynamics, respectively. As a result, the time complexity of \ours\ is expressed as $l\cdot O(GCN)+O(GRNN)$, where $l$ indicates the number of timestamps (\ie, the number of KG snapshots). Specifically, the time complexity of \texttt{GCN} is defined as $O(L\cdot (n+j+d(d+1))$~\cite{Zonghan19}, where $L$, $n$, $j$, and $d$ represent the number of \texttt{GCN} layers, the number of nodes (\ie, users and companies), the number of edges (\ie, positions), and the dimensionality of embeddings, respectively. In addition, the time complexity of \texttt{GRNN} is given by $O(l\cdot ({d_h}(d_h+1)+ d_i))$~\cite{RotmanW21}, where $d_i$ and $d_h$ indicate the input and hidden size of RNN, respectively. 

Here, one might have concerns about the computational overhead of \ours.
Considering the architecture that repeats \texttt{GCN} and \texttt{GRNN} $l$ times, the value of $l$ can impact the performance of \ours. 
However, we observed two key points: (1) the average inference time per user is only 0.89 msec with \ours, suggesting that it provides real-time predictions, and (2) \ours\ shows linear scalability as the number of users increases.
Moreover, we can mitigate the computational overhead of \ours\ on large datasets without significant accuracy loss by reducing the value of $l$ for training each temporal embedding, as validated in the \href{https://github.com/Bigdasgit/CAPER/blob/main/Online_Appendix_CAPER_.pdf}{{\ul{online appendix}}}.

\section{Evaluation}\label{sec:evaluation}
To evaluate effectiveness of \ours, we designed our experiments, aiming at answering the following key evaluation questions (EQs):
\begin{itemize}[leftmargin=*]
    \item {\textbf{(EQ1})} Does \ours\ outperform its competitors for career trajectory prediction?
    \item {\textbf{(EQ2})} Is temporal knowledge graph modeling effective for career trajectory prediction?
    \item {\textbf{(EQ3})} Is learning mutual dependency effective for career trajectory prediction?
    \item {\textbf{(EQ4})} Is learning characteristic shifts over time effective for career trajectory prediction?
\end{itemize}
We also demonstrated that 
the accuracies of \ours\ do 
not fluctuate a lot depending on the embedding dimensionality, and that \ours\ works well in a different time granularity 
(see \href{https://github.com/Bigdasgit/CAPER/blob/main/Online_Appendix_CAPER_.pdf}{{\ul{online appendix}}}).

\subsection{Experimental Settings}\label{sec:eval_setting}

\noindent{\bf Dataset.} 
Table~\ref{table:dataset} provides some statistics on our dataset.
We used a real-world career trajectory dataset whose
 details were explained in Section~\ref{sec:pre_data}. 
For the extrapolated career reasoning task 
(\textit{additionally designed by us}) 
discussed in Section~\ref{sec:sec-overview}, we used the most recent five years of careers (\ie, $M$=5) for the test set and the remaining 47 years for the training set.
It is worth noting that finding publicly available datasets is very challenging in a CTP domain due to the proprietary nature of the datasets. 
Privacy concerns surrounding resume datasets
further hinder their public release.
Therefore, most prior literature, including NEMO \cite{li2017nemo}, HCPNN \cite{meng2019hierarchical}, AHEAD \cite{zhang2021attentive}, TACTP \cite{wang2021variable}, and NAOMI \cite{yamashita2022looking}, 
used a single private dataset of their own, none of which was shared publicly. 
For this reason, we had to rely only on our own dataset (obtained from a global career platform, FutureFit AI) for the evaluation.

For the benefit of the research community, 
our dataset is available upon request; interested researchers should contact FutureFit AI or the Penn State team to access the original dataset.
Additionally, in~\cite{Yamashita24}, we introduced OpenResume, a publicly available, anonymized, and structured resume dataset, specifically curated for downstream tasks in the job domain.

\vspace{1mm}
\noindent{\bf Competitors.} 
To evaluate the effectiveness of \ours, we compare \ours\ with four baselines, two recent TKG reasoning methods, and five state-of-the-art CTP methods
-- (1) baselines: Popular (\ie, predicts the most-frequent companies and positions), RNN, GRU, and Transformer~\cite{VaswaniSPUJGKP17}; 
(2) TKG reasoning methods: DaeMon~\cite{DongNWQWZF23} and CENET~\cite{XuO0F23};
(3) CTP methods: NEMO \cite{li2017nemo}, HCPNN \cite{meng2019hierarchical}, AHEAD \cite{zhang2021attentive}, TACTP \cite{wang2021variable}, and NAOMI \cite{yamashita2022looking}.

As we have carefully surveyed the relevant literature, including~\cite{Qin23survey}, it is important to note that, to the best of our knowledge, our competitors here cover all the relevant baselines in the CTP problem (refer to the `Career Mobility Prediction' category in~\cite{Qin23survey}).
For evaluation, we used the source codes provided by the authors.
Since none of the competitors was designed for multi-step inference, we call those algorithms repeatedly to predict the next companies and positions.
We carefully tuned the hyperparameters of competitors and \ours.
For the full reproducibility, we provide complete implementation details in the Appendix.

\vspace{1mm}
\noindent{\bf Evaluation Tasks.} 
Following~\cite{zhang2021attentive, wang2021variable}, we employ company and position prediction tasks.
The goal of both tasks is to forecast each user's next careers (\ie, company or position) that are likely to occur at \emph{non-existent future $M$ timestamps} in the training set. 
Specifically, we consider only those users (\ie, having their careers after 2015) included in the test set as test users.
Given each test user's career trajectory during the training period, we predict the next companies and positions that the user is likely to work in the test timestamps.
Since we are already addressing the problem of long-term career prediction, we do not additionally perform the duration prediction task done in~\cite{meng2019hierarchical, zhang2021attentive, wang2021variable}, which only predicts the single next career.

Following \cite{li2017nemo, meng2019hierarchical, zhang2021attentive, wang2021variable}, we employ the two popular metrics: mean reciprocal rank (MRR) and Accuracy (Acc) @$k$ ($k$=1, 5, 10). 
Using these metrics, we measure how accurately each method predicts each test user's company or position per test timestamp (\ie, from 2016 to 2020). 
We train each method over 3 random weight initializations and report the averaged accuracy.

\vspace{1mm}
\noindent{\bf Iadequateness of Random-Sampling-Based Evaluation.} We note that some CTP methods~\cite{zhang2021attentive,wang2021variable} 
evaluate their CTP solutions as follows: they randomly sample 10\% of users and use their career trajectories to construct a test set.
However, we believe this evaluation to be problematic.

\begin{figure}[t]
\centering
  \includegraphics[width=0.97\linewidth]{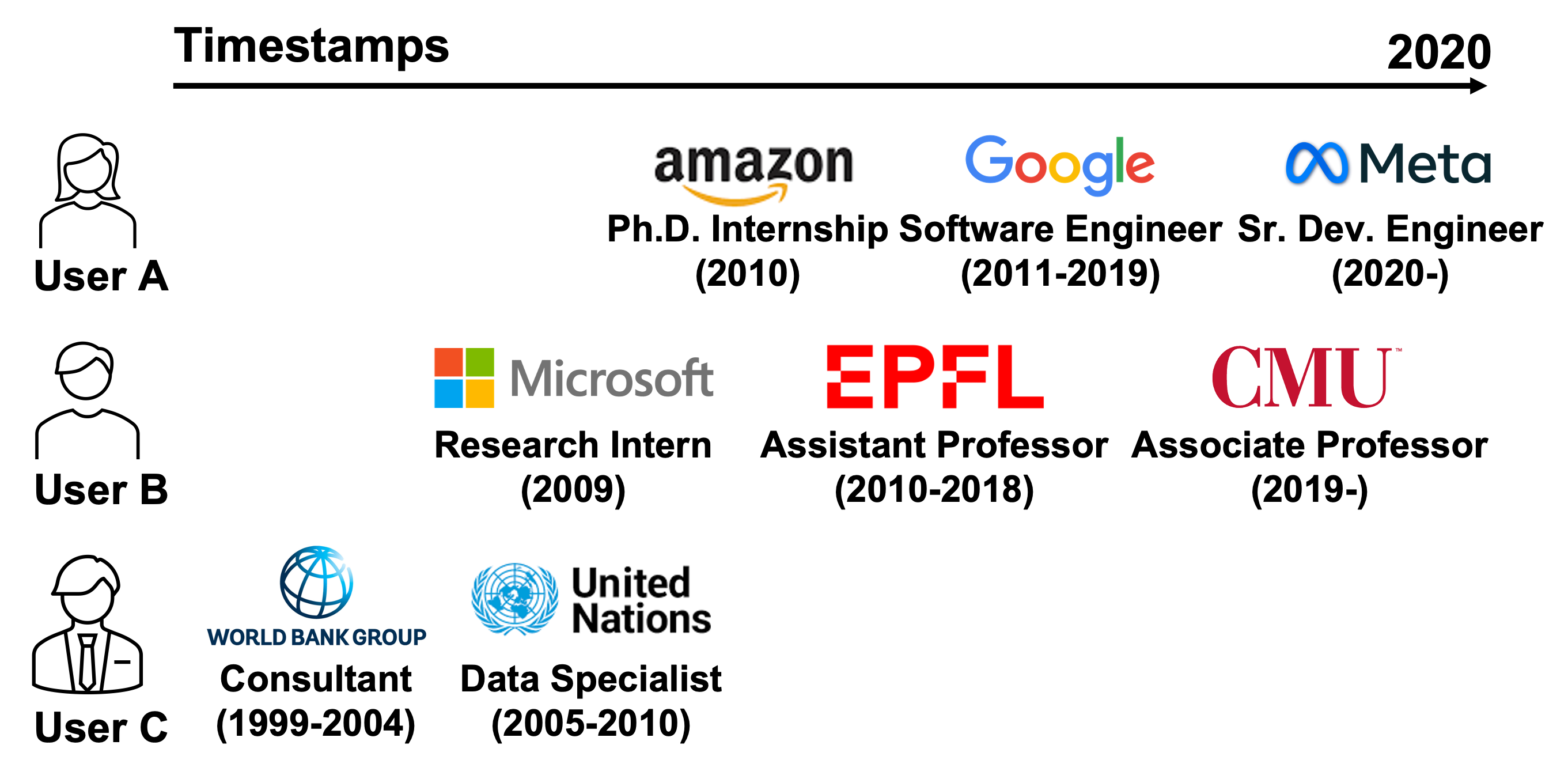} 
\caption{An example to show the limitations of the random-sampling-based evaluation employed in existing studies.} \label{fig:task}
\vspace{-0.2cm}
\end{figure}

Consider Figure~\ref{fig:task} that depicts the career trajectories of three users.
Suppose that the career trajectories of two users, A and B, are used for the training set, while another user C's career trajectory is used for the test set.
In this case, where test user C has a career trajectory from 1999 to 2010, a CTP method should predict careers that occur at existent past timestamps in the training set. 
Specifically, since test user C has his/her last career at 2010 for some reason (\eg, retirement), he/she doesn't need predictions for the next career anymore.
We believe such an evaluation is not realistic in terms of real-world applications (\eg, future career prediction).
We found that among all 333K users on our dataset (see Table~\ref{table:dataset}), the proportion of users whose careers do not exist at the test timestamps (like user C) is 55.06\% (\ie, 149K users),
which indicates they can be prone to be inappropriately sampled as test users under the random-sampling-based evaluation.
To address this limitation, we consider only users (\eg, A and B in Figure~\ref{fig:task}) who have careers at the test timestamps (\ie, from 2016 to 2020) as test users, and their careers (\eg, Google, Meta, EPFL, and Carnegie Melon University in Figure~\ref{fig:task}) as the ground truth for evaluation.

\begin{table*}[!t]
\centering
\caption{Accuracies of four baselines, two recent TKG reasoning methods, and five state-of-the-art CTP methods, and \ours. We conducted \boldsymbol{$t$}-tests at a 95\% confidence level, showing statistically significant differences between \ours\ and competitors, with \boldsymbol{$p$}-values below 0.05 (except for \boldsymbol{$M$}=4 on Acc@10 in Table~\ref{table:comparison}-(a)).
}
\vspace{-0.2cm}
\label{table:comparison}
\resizebox{\textwidth}{!} {
 \renewcommand{\arraystretch}{1.4}
\begin{tabular}{c|ccccc|ccccc|ccccc|ccccc}
\multicolumn{21}{c}{\LARGE\textbf{(a) Company Prediction}} \\ \toprule
\multicolumn{1}{c|}{\multirow{2}{*}{\textbf{Methods}}} & \multicolumn{5}{c|}{\textbf{MRR}} & \multicolumn{5}{c|}{\textbf{Acc@1}} & \multicolumn{5}{c|}{\textbf{Acc@5}} & \multicolumn{5}{c}{\textbf{Acc@10}}\\ 
 & \textbf{\boldsymbol{$M$}=1} & \textbf{\boldsymbol{$M$}=2} & \textbf{\boldsymbol{$M$}=3} & \textbf{\boldsymbol{$M$}=4} & \textbf{\boldsymbol{$M$}=5} & \textbf{\boldsymbol{$M$}=1} & \textbf{\boldsymbol{$M$}=2} & \textbf{\boldsymbol{$M$}=3} & \textbf{\boldsymbol{$M$}=4} & \textbf{\boldsymbol{$M$}=5} & \textbf{\boldsymbol{$M$}=1} & \textbf{\boldsymbol{$M$}=2} & \textbf{\boldsymbol{$M$}=3} & \textbf{\boldsymbol{$M$}=4} & \textbf{\boldsymbol{$M$}=5} & \textbf{\boldsymbol{$M$}=1} & \textbf{\boldsymbol{$M$}=2} & \textbf{\boldsymbol{$M$}=3} & \textbf{\boldsymbol{$M$}=4} & \textbf{\boldsymbol{$M$}=5} \\ \midrule
 \textbf{Popular} &	1.51	&	1.47	&	1.45	&	1.49	&	1.5
&	0.42	&	0.38	&	0.33	&	0.34	&	0.34
&	0.90	&	0.85	&	0.80	&	0.84	&	0.85
&	3.12	&	3.06	&	3.05	&	3.16	&	3.18
\\ 
\textbf{RNN} &	37.32	&	24.17	&	19.15	&	17.32	&	16.42	&	32.29	&	20.52	&	16.15	&	14.59	&	13.91	&	41.93	&	26.56	&	20.88	&	18.87	&	17.77	&	46.42	&	30.82	&	24.56	&	22.20	&	20.82

 \\ 
 \textbf{GRU} &	40.90	&	27.05	&	21.76	&	19.84	&	18.99	&	36.11	&	23.57	&	18.77	&	17.07	&	16.43	&	45.30	&	29.26	&	23.58	&	21.48	&	20.50	&	49.55	&	33.38	&	27.12	&	24.79	&	23.53

 \\ 
\textbf{Transformer} &	38.88 	&	32.30	&	28.77	& 26.42	&	25.07	&	27.96	&	24.92	&	21.46	&	17.47	&	15.17	&	45.46	&	34.57	&	30.18	&	29.63	&	28.94	&	50.23	&	38.62	&	34.73	&	34.18	&	34.10

 \\ \midrule
\textbf{DaeMon} &	53.70	&	42.51	&	38.24	&	35.16	&	30.25	&	44.32	&	35.24	&	25.26	&	21.20	&	18.57	&	59.83	&	40.25	&	35.22	&	30.88	&	26.80	&	\ul{64.27}	&	50.20	&	36.90	&	32.48	&	29.51
  \\ 
\textbf{CENET} &	\ul{60.03}	&	47.52	&	42.74	& 39.30	&	33.81	&	56.39	&	44.83	&	32.13	&	26.97	&	23.62	&	\ul{62.96}	&	42.35	&	37.06	&	32.49	&	28.20	&	63.09	&	49.27	&	36.20	&	31.88	&	28.96
  \\ \midrule 
\textbf{NEMO} &	41.99	&	31.19	&	26.24	&	24.00	&	22.46	&	37.17	&	27.31	&	22.77	&	20.69	&	19.26	&	47.54	&	34.92	&	29.43	&	26.97	&	25.29	&	50.57	&	38.08	&	32.32	&	29.82	&	28.08
  \\ 
\textbf{HCPNN} &	44.55	&	35.02	&	30.97	&	29.57	&	29.06	&	40.93	&	32.51	&	28.60	&	27.24	&	26.80	&	48.18	&	37.15	&	32.93	&	31.47	&	30.88	&	50.76	&	39.32	&	35.07	&	33.61	&	32.95
  \\ 
\textbf{AHEAD} &	44.72	&	33.71	&	29.91	&	28.66	&	28.15	&	40.60	&	30.93	&	27.29	&	26.04	&	25.55	&	49.39	&	36.04	&	32.02	&	30.75	&	30.17	&	52.04	&	38.50	&	34.44	&	33.24	&	32.70  \\ 
\textbf{NAOMI} & 44.22 & 	43.58 & 	41.75 & 	42.13 & 	43.20 & 	38.53 & 	38.10 & 	36.33 & 	36.37 & 	37.37	 & 48.64	 & 48.62 & 	44.99 & 46.14	 & 47.91 & 	50.17 & 	48.29	 & 46.33 & 	46.98 & 	49.10

  \\
\textbf{TACTP} & 59.08	& \ul{54.38}	& \ul{49.83}	& \ul{48.70}	& \ul{48.18}	& \ul{57.57}	& \ul{52.95}	& \ul{49.12}	& \ul{47.24}	& \ul{46.89}	& 60.23	& \ul{55.25}	& \ul{51.12}	& \ul{49.91}	& \ul{48.68}	& 61.10	& \ul{56.58}	& \ul{51.99}	& \bf{51.70}	& \ul{49.16} \\
 \midrule
\textbf{\ours} &	\bf{71.07}	&	\bf{58.14}	&	\bf{51.35}	&	\bf{49.70}	&	\bf{49.43}	&	\bf{69.34}	&	\bf{57.09}	&	\bf{49.44}	&	\bf{48.24}	&	\bf{48.58}	&	\bf{73.27}	&	\bf{59.05}	&	\bf{51.75}	&	\bf{50.75}	&	\bf{50.09}	&	\bf{74.10}	&	\bf{59.93}	&	\bf{52.87}	&	\ul{51.57}	&	\bf{50.88}
\\
% \textbf{\boldsymbol{$p$}-value} & \bf{2.21e-3}	
% & \bf{9.84e-3} & \bf{1.60e-2} & \bf{2.15e-2} & \bf{8.96e-3} & \bf{2.35e-3} & \bf{1.38e-2} & \bf{4.16e-2} & \bf{1.56e-2} & \bf{9.62e-3} & \bf{2.04e-3} & \bf{1.02e-2} & \bf{3.62e-2} & \bf{2.36e-2} & \bf{3.09e-2} & \bf{2.24e-3} & \bf{8.53e-3} & \bf{1.78e-2} & \bf{6.80e-2} & \bf{1.90e-2}
%  \\ 
 \bottomrule
 \addlinespace[1em]
 \multicolumn{21}{c}{\LARGE\textbf{(b) Position Prediction}} \\ \toprule
\multicolumn{1}{c|}{\multirow{2}{*}{\textbf{Methods}}} & \multicolumn{5}{c|}{\textbf{MRR}} & \multicolumn{5}{c|}{\textbf{Acc@1}} & \multicolumn{5}{c|}{\textbf{Acc@5}} & \multicolumn{5}{c}{\textbf{Acc@10}}\\ 
& \textbf{\boldsymbol{$M$}=1} & \textbf{\boldsymbol{$M$}=2} & \textbf{\boldsymbol{$M$}=3} & \textbf{\boldsymbol{$M$}=4} & \textbf{\boldsymbol{$M$}=5} & \textbf{\boldsymbol{$M$}=1} & \textbf{\boldsymbol{$M$}=2} & \textbf{\boldsymbol{$M$}=3} & \textbf{\boldsymbol{$M$}=4} & \textbf{\boldsymbol{$M$}=5} & \textbf{\boldsymbol{$M$}=1} & \textbf{\boldsymbol{$M$}=2} & \textbf{\boldsymbol{$M$}=3} & \textbf{\boldsymbol{$M$}=4} & \textbf{\boldsymbol{$M$}=5} & \textbf{\boldsymbol{$M$}=1} & \textbf{\boldsymbol{$M$}=2} & \textbf{\boldsymbol{$M$}=3} & \textbf{\boldsymbol{$M$}=4} & \textbf{\boldsymbol{$M$}=5} \\ \midrule
  \textbf{Popular} &	10.66	&	10.64	&	10.54	&	10.45	&	10.43
&	3.51	&	3.29	&	2.99	&	2.87	&	2.89
&	8.82	&	8.94	&	8.88	&	8.73	&	8.73
&	25.12	&	25.61	&	25.99	&	26.06	&	25.92
 \\ 
\textbf{RNN} &	27.41	&	18.90	&	15.79	&	14.31	&	13.47	&	19.29	&	11.04	&	8.06	&	6.35	&	5.74	&	32.61	&	23.69	&	20.24	&	19.07	&	17.44	&	45.43	&	34.45	&	31.35	&	30.41	&	28.83
 \\ 
\textbf{GRU} &	30.05	&	20.66	&	18.32	&	17.57	&	17.72	&	22.05	&	12.62	&	10.40	&	9.41	&	9.53	&	34.79	&	25.49	&	22.99	&	22.71	&	22.60	&	47.92	&	36.57	&	34.18	&	33.87	&	33.50
 \\ 
 \textbf{Transformer} &	28.83	&	26.64	&	22.05	& 21.92	&	20.75	&	20.13	&	18.89	&	13.31	&	12.76	&	11.63	&	34.01	&	29.81	&	25.32	&	22.36	&	21.28	&	48.13	&	44.07	&	40.08	&	36.47	&	33.93

 \\ \midrule
\textbf{DaeMon} &	40.85	&	33.84	&	21.54	&	19.60	&	17.54	&	31.50	&	18.88	&	13.54	&	12.74	&	11.44	&	51.40	&	44.63	&	35.12	&	34.76	&	31.58	&	59.38	&	50.50	&	42.67	&	41.00	&	40.51
  \\ 
\textbf{CENET} &	\ul{51.08}	&	\ul{42.31}	&	26.93	& 24.51	&	21.93	&	\ul{45.46}	&	27.25  &	19.54	&	18.38	&	16.51	&	\ul{54.89}	&	\ul{47.66}	&	37.50	&	37.12	&	33.72	&	\ul{60.44}	&	51.40	&	43.43	&	41.73	&	41.23
  \\ \midrule 
\textbf{NEMO} &	34.32	&	23.22	&	20.11	&	18.43	&	17.95	&	25.49	&	13.82	&	11.08	&	9.24	&	8.97	&	42.96	&	32.00	&	28.23	&	26.89	&	25.57	&	53.45	&	42.20	&	38.82	&	37.60	&	36.42
  \\ 
\textbf{HCPNN} &	11.39	&	11.20	&	11.10	&	11.08	&	10.95	&	4.98	&	4.19	&	4.15	&	4.14	&	4.08	&	15.84	&	14.75	&	14.62	&	14.60	&	14.40	&	26.04	&	24.87	&	24.59	&	24.52	&	24.24  \\ 
\textbf{AHEAD} &	33.59	&	25.17	&	22.66	&	21.26	&	20.98	&	24.58	&	15.62	&	13.06	&	11.27	&	11.07	&	42.07	&	34.09	&	31.54	&	30.91	&	30.07	&	53.42	&	44.32	&	42.33	&	41.88	&	41.14  \\ 
\textbf{NAOMI}  & 30.81	 & 30.33	 & 30.24	 & 30.13 & 	29.48 & 	23.61 & 	23.34	 & 23.38 & 23.30 & 	22.99 & 	40.64	 & 40.68	 & 36.99	 & 38.14 & 	39.91	 & 48.17 & 	46.29	 & 44.33 & 	44.98 & 	47.10

 \\
\textbf{TACTP} & 37.09	& 36.14	& \ul{35.00}	& \ul{34.33}	& \ul{33.34}	& 28.73	& \ul{27.71}	& \ul{26.84}	& \ul{25.99}	& \ul{24.82}	& 43.53	& 42.78	& \ul{41.00}	& \ul{40.08}	& \ul{39.69}	& 52.91	& \ul{51.97}	& \ul{50.89}	& \ul{50.64}	& \ul{48.51} \\
 \midrule
\textbf{\ours} & \bf{59.92}	&	\bf{48.66}	&	\bf{43.28}	&	\bf{43.15}	&	\bf{44.03}	&	\bf{50.87}	&	\bf{40.62}	&	\bf{35.74}	&	\bf{35.62}	&	\bf{36.63}	&	\bf{69.69}	&	\bf{56.56}	&	\bf{50.27}	&	\bf{50.42}	&	\bf{50.77}	&	\bf{76.35}	&	\bf{62.47}	&	\bf{56.61}	&	\bf{56.53}	&	\bf{56.58}
\\
\bottomrule
\end{tabular}
}
\vspace{-0.1cm}
\end{table*}

\subsection{Results}\label{sec:eval_results}
For EQ2, EQ3, and EQ4, we only show the results of \ours\ when the value of $M$ is 1 (\ie, career prediction for the immediate next year).
However, we confirmed that the performance for M=[2-5] is consistent with that for M=1 (see \href{https://github.com/Bigdasgit/CAPER/blob/main/Online_Appendix_CAPER_.pdf}{{\ul{online appendix}}}).

\vspace{1mm}
\noindent{\bf (EQ1) Comparison with 11 Competitors.} 
We conducted comparative experiments on our career trajectory dataset.
to demonstrate the superiority of \ours\ over the four baselines, two TKG reasoning methods, and five state-of-the-art CTP methods.
In Table~\ref{table:comparison},
the values in \textbf{boldface} and \ul{underlined} indicate the best and 2nd best accuracies in each column, respectively. 

From Table~\ref{table:comparison}, we can see that \ours\ consistently and significantly outperforms all competitors in almost all cases.
% (except for Acc@5, 10 of the company prediction when $M$ is 4). 
Overall, \ours\ yields 6.80\% and 34.58\% more-accurate company and position predictions on average, respectively, compared to the best competitor (\ie, TACTP).
Specifically, for company prediction, \ours\ yields up to 20.29\% and 6.91\% higher MRR than the best competitor when the values of $M$ are 1 and 2 (\ie, short-term predictions, which are the most practical scenario in our setting), respectively.
For long-term predictions (\ie, $M$=3,4,5), \ours\ also provides accuracy slightly better than or comparable to the best competitor.
Additionally, for position prediction, \ours\ 
yields up to 61.55\%, 34.64\%, 23.65\%, 25.69\%, and 32.06\% higher MRR than the best competitor when the values of $M$ are 1, 2, 3, 4, and 5, respectively. 
We attribute the superiority of \ours\ to its sophisticated career modeling and capturing of characteristic shifts of entities over time.
That is, it validates our claims for the significance of considering mutual dependency and characteristic shifts in the CTP problem, as discussed in Section~\ref{sec:discussion}.

\begin{table*}[!t]
\centering
\caption{Ablation studies for \ours\ when the value of \boldsymbol{$M$} is 1. The goals of (EQ2), (EQ3), and (EQ4) are to demonstrate the effectiveness of (M1), (M2), and (M3) in \ours, respectively. 
The results show that not only each design choice in \ours\ is effective, but also \ours\ integrating all of them is the most effective.
}
\vspace{-0.25cm}
\label{table:ablation}
\resizebox{0.85\textwidth}{!} {
\begin{tabular}{c|cc|c|cccc|cccc||c}
\multicolumn{13}{c}{\textbf{(a) Company Prediction}} \\ \toprule
\multicolumn{1}{c|}{\multirow{2}{*}{\textbf{Metrics}}} & \multicolumn{2}{c|}{\textbf{(\rom{1}) EQ2}} & \multicolumn{1}{c|}{\textbf{(\rom{2}) EQ3}} & \multicolumn{4}{c|}{\textbf{(\rom{3}) EQ4-1}} & \multicolumn{4}{c||}{\textbf{(\rom{4}) EQ4-2}} & \multirow{2}{*}{\textbf{\ours}} \\ 
& \textbf{KG} & \textbf{HetG} & \textbf{w/o GCN} & \textbf{All} & \textbf{w/o $\tilde{\mathbf{u}}_x$} & \textbf{w/o $\tilde{\mathbf{c}}_x$} & \textbf{w/o $\tilde{\mathbf{p}}_x$} & \textbf{w/o GRNN}  & \textbf{RNN} & \textbf{GRU} & \textbf{Transformer} & \\ \midrule
\textbf{MRR} & 64.40	&	61.54	&	7.44	&	69.94	&	55.36	&	69.17	&	\textbf{71.07}	&	56.01	&	69.99	&	69.83	&	66.87
 & \textbf{71.07} \\ 
\textbf{Acc@1} & 60.64	&	58.11	&	4.65	&	68.67	&	51.50 &	67.76	&	\textbf{69.34}	&	53.48	&	68.20	&	68.08	&	63.71
 & \textbf{69.34} \\ 
\textbf{Acc@5}& 66.28	&	65.21	&	9.41	&	70.51	&	59.54	&	69.78	&	\textbf{73.27}	&	56.97	&	70.80	&	70.70	&	70.28
 & \textbf{73.27} \\
\textbf{Acc@10} & 71.35	&	67.86	&	12.49	&	72.10	&	62.55	&	71.61	&	\textbf{74.10}	&	60.69	&	73.12	&	72.87	&	72.53
 & \textbf{74.10} \\ \bottomrule
\addlinespace[1em]
\multicolumn{13}{c}{\textbf{(b) Position Prediction}} \\ \toprule
\multicolumn{1}{c|}{\multirow{2}{*}{\textbf{Metrics}}} & \multicolumn{2}{c|}{\textbf{(\rom{1}) EQ2}} & \multicolumn{1}{c|}{\textbf{(\rom{2}) EQ3}} & \multicolumn{4}{c|}{\textbf{(\rom{3}) EQ4-1}} & \multicolumn{4}{c||}{\textbf{(\rom{4}) EQ4-2}} & \multirow{2}{*}{\textbf{\ours}} \\ 
& \textbf{KG} & \textbf{HetG} & \textbf{w/o GCN} & \textbf{All} & \textbf{w/o $\tilde{\mathbf{u}}_x$} & \textbf{w/o $\tilde{\mathbf{c}}_x$} & \textbf{w/o $\tilde{\mathbf{p}}_x$} & \textbf{w/o GRNN}  & \textbf{RNN} & \textbf{GRU} & \textbf{Transformer} & \\ \midrule
\textbf{MRR} & 46.47	&	45.75	&	3.85	&	16.70	&	40.40	&	17.74	&	\textbf{59.92}	&	30.37	&	58.95	&	58.01	&	52.74
 & \textbf{59.92} \\ 
\textbf{Acc@1} & 28.82	&	29.07	&	0.06	&	7.94	&	27.74	&	8.66	&	\textbf{49.87}	&	21.21	&	49.77	&	48.00	&	40.79
 & \textbf{49.87} \\ 
\textbf{Acc@5} & 60.41	&	66.95	&	2.65	&	22.96	&	55.15	&	24.71	&	\textbf{69.69}	&	32.91	&	64.71	&	64.50	&	66.40
 & \textbf{69.69} \\
\textbf{Acc@10} & 75.73	&	73.5	&	9.83 &	34.74 &	63.39	&	36.59	&	\textbf{76.35}	&	48.23	&	74.09	&	74.67	&	72.93
 & \textbf{76.35} \\ 
 \bottomrule
\end{tabular}
}
\vspace{-0.2cm}
\end{table*}

\vspace{1mm}
\noindent{\bf (EQ2) Effectiveness of TKG Modeling.} 
In (M1) of Section~\ref{sec:sec-modules}, we modeled the given career trajectories as a TKG to represent the mutual dependency between key units and their characteristic shifts over time. 
For (EQ2), we compare \ours\ with its two variants: \ours$_{\textbf{KG}}$ models the career trajectories as a KG without consideration of temporal dimension; and \ours$_{\textbf{HetG}}$ models the career trajectories as a heterogenous graph, where nodes indicate companies and positions, and edges indicate the career transitions between two companies or two positions and whether a position belongs to a company~\cite{zhang2021attentive}. 
As shown in Table~\ref{table:ablation}-(\rom{1}), we see that \ours\ consistently outperforms its two variants. 
For company (resp. position) prediction, \ours\ yields up to 10.35\% and 15.48\% (resp. 28.94\% and 30.97\%) higher MRR than \ours$_{\textbf{KG}}$ and \ours$_{\textbf{HetG}}$, respectively.
For position prediction, \ours\ yields up to 28.94\% and 30.97\% higher MRR than \ours$_{\textbf{KG}}$ and \ours$_{\textbf{HetG}}$, respectively.
It shows that not only preserving the mutual dependency for the ternary relationships on users, positions, and companies (\ie, \ours\ and \ours$_{\textbf{KG}}$ vs. \ours$_{\textbf{HetG}}$) but also adding a temporal dimension to KG (\ie, \ours\ vs. \ours$_{\textbf{KG}}$) is helpful in better modeling career trajectories.

\vspace{1mm}
\noindent{\bf (EQ3) Effectiveness of Learning Mutual Dependency.} 
In (M2) of Section~\ref{sec:sec-modules}, we designed a GCN encoder to learn mutual dependencies for the ternary relationships of a user, a position, and a company.
For (EQ3), we compare \ours\ with its variant \ours$_{\textbf{w/o GCN}}$ (\ie, not employing a GCN encoder and
forwarding randomly initialized temporal embeddings into the GRNN encoder without aggregating  career information).
Table~\ref{table:ablation}-(\rom{2}) shows that \ours\ dramatically outperforms \ours$_{\textbf{w/o GCN}}$. 
The results show that our convolutional layer benefits from learning mutual dependency for ternary relationships.

\vspace{1mm}
\noindent{\bf (EQ4) Effectiveness of Learning Characteristic Shifts.} 
In (M3) of Section~\ref{sec:sec-modules}, we represented the same unit (\spec, user and company) at different timestamps as different temporal embeddings. 
Based on the temporal embeddings, we then reflected the changes in the characteristics of users and companies in their evolution embeddings gradually.
To verify the effectiveness of both ideas, 
we conduct experiments to answer the two sub-questions:
\begin{itemize}[leftmargin=*]
    \item {\textbf{(EQ4-1)}} Is it effective to consider characteristic shifts of all units?
    \item {\textbf{(EQ4-2)}} Is it effective to model the evolution over time of units  based on GRNN?
\end{itemize}

For (EQ4-1), we compare the accuracy of the following 4 variants of \ours:
(1) \ours$_{\textbf{all}}$ represents all units 
at different timestamps as different temporal embeddings; (2) \ours$_{\textbf{w/o } \tilde{\mathbf{u}}_x}$ represents each user as a single embedding regardless of timestamps; (3) \ours$_{\textbf{w/o } \tilde{\mathbf{c}}_x}$ represents each company as a single embedding regardless of timestamps; (4) \ours$_{\textbf{w/o } \tilde{\mathbf{p}}_x}$ represents each position as a single embedding regardless of timestamps. 
To implement \ours$_{\textbf{all}}$, we add another GRNN encoder for positions in Figure~\ref{fig:overview}.
Here, \textbf{\ours}\boldsymbol{$_{\textbf{w/o } \tilde{\mathbf{p}}_x}$} \textbf{is the same as the final version of \ours}.

As shown in Table~\ref{table:ablation}-(\rom{3}), \ours$_{\textbf{w/o } \tilde{\mathbf{p}}_x}$ consistently outperforms other variants of \ours.
Notably, we see that for position prediction, using evolution embeddings of positions adversely affects the performance of \ours.
On the other hand, we note that jointly considering the characteristic shifts of users and companies is consistently helpful for solving CTP problem. 
We conjecture that the reason of this phenomenon is that, unlike users and companies, the characteristics of positions tend to have a more-static nature that does not change significantly over time. 
For instance, the nature of ``software engineer" position remains more static between 2010 and 2020 (despite new positions getting created) while the skills of users or characteristics of companies change more dynamically. 

For (EQ4-2), we compare \ours\ (\ie, \ours$_{\textbf{LSTM}}$) with its variant \ours$_{\textbf{w/o GRNN}}$, which does not employ any GRNN encoders and thus cannot construct evolution embeddings of users and companies.
Table~\ref{table:ablation}-(\rom{4}) illustrates the importance of learning the evolution of users and companies over time by sequentially accumulating their past career information.
Additionally, we found that employing RNN or GRU encoders yields comparable results.
Conversely, we observed that the Transformer encoder exhibits limited effectiveness in terms of accuracy in the CTP domain.
This suggests that the sequential order of each user's career is particularly important in this domain.
From the results, we believe that our superior performance can be attributed to our novel modeling and utilization of CTP data, rather than specific techniques.

\section{Conclusions}\label{sec:conclusions}
In this work, we found that
existing CTP methods
fail to jointly consider the mutual dependency for ternary relationships among a user, a position, and a company, and the characteristic shifts over time.
To address the challenges, we proposed a novel CTP approach, \ours, that consists of three key modules: (M1) modeling career trajectories, (M2) learning mutual dependency, and (M3) learning temporal dynamics.
Furthermore, we formulated the CTP problem as the extrapolated career reasoning task on TKG, which aims to forecast the next careers that are likely to occur at non-existent future timestamps in the input TKG. 
Via experiments using a real-world career trajectory dataset, we demonstrated that 
\ours\ consistently and significantly outperforms 4 baselines, 2 recent TKG reasoning methods, and 5 state-of-the-art CTP methods.

\begin{acks}
The work of Sang-Wook Kim was supported by the Institute of Information \& communications Technology Planning \& Evaluation (IITP) grant funded by the Korea government(MSIT) (No.2022-0-00352 and No.RS-2022-00155586, A High-Performance Big-Hypergraph Mining Platform for Real-World Downstream Tasks).
Yeon-Chang Lee's work was supported by the Institute of Information \& communications Technology Planning \& Evaluation (IITP) grant, funded by the Korea government (MSIT) (No. RS-2020-II201336, Artificial Intelligence Graduate School Program (UNIST)).
Dongwon Lee's was supported by NSF award \#1934782.
The dataset was provided by FutureFit AI under a Memorandum of Understanding.
\end{acks}

\bibliographystyle{ACM-Reference-Format}
\balance
\bibliography{bibliography}

\renewcommand\thesection{\Alph{section}}
\renewcommand\thetable{\Roman{table}}
\renewcommand\thefigure{\Roman{figure}}
\setcounter{section}{0}
\setcounter{table}{0}
\setcounter{figure}{0}

\section{Further Data Description}\label{sec:dataset}

We pre-process and standardize job titles (\ie, position names) and company names.
For example, the position of ``software engineer" may be written as ``software development engineer" or ``SDE" in different resumes for various reasons (\eg, lack of industry standard taxonomy across companies or individual preference). 
However, in order to successfully solve the CTP problem, we  need to identify and coalesce all differently-worded position names that actually refer to the same or very similar positions.
For obtaining such normalized position names, we performed entity mapping based on the state-of-the-art job title mapping method \cite{yamashita2023james}, which cleans job titles into the ECSO-accepted occupations
(\ie, the public job taxonomy). 
In the end, the position names in our dataset are mapped to 425 normalized position names. 
For cleaning company names, following the previous works \cite{li2017nemo, zhang2021attentive}, 
we use the most frequently-occurred $9,000$ companies while filtering out low frequent companies.

\begin{figure}[t]
    \centering
    \subfloat[\footnotesize Normalized position name]{\label{b}\includegraphics[width=.44\linewidth]{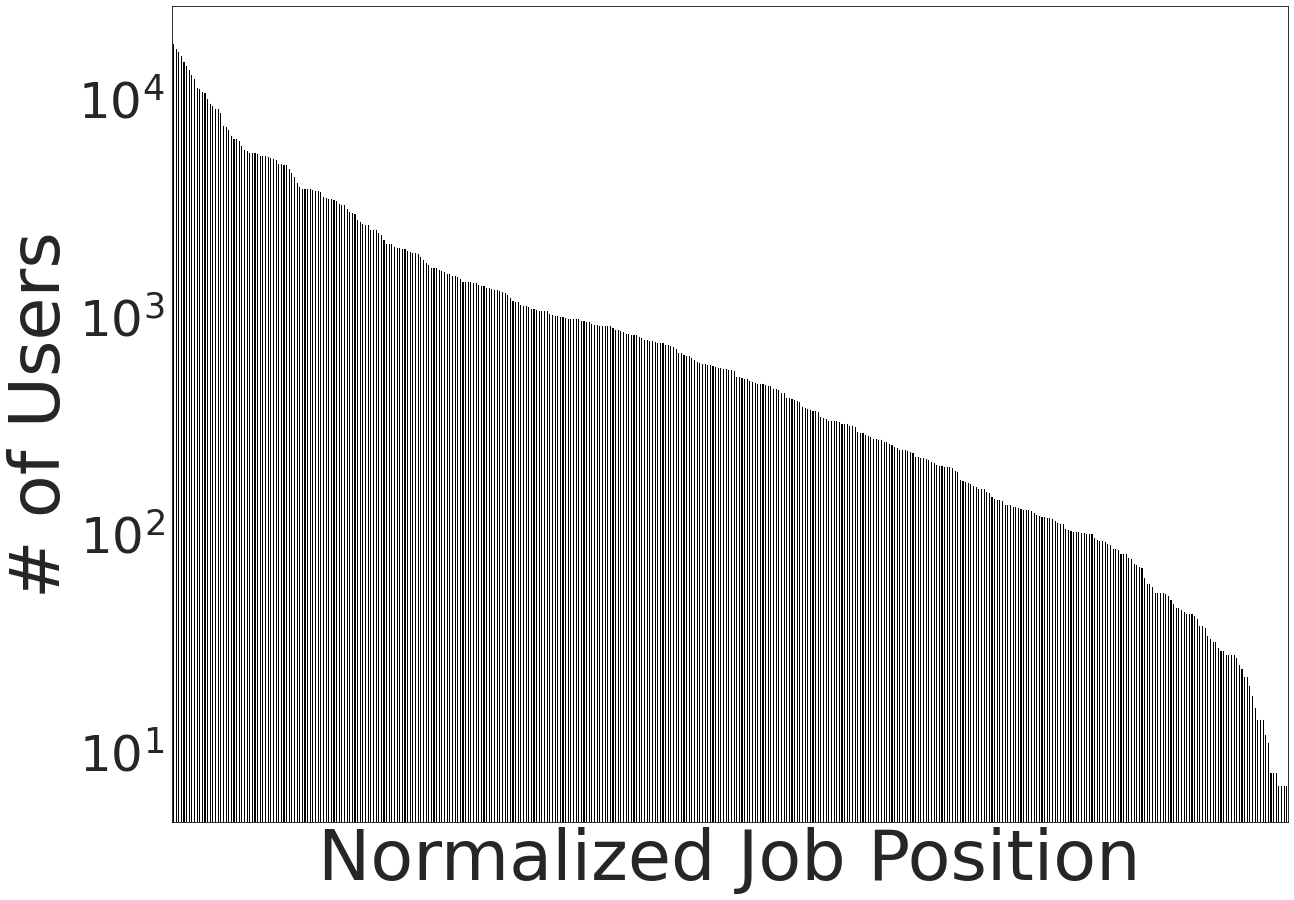}}
    \subfloat[\footnotesize Company name]{\label{c}\includegraphics[width=.44\linewidth]{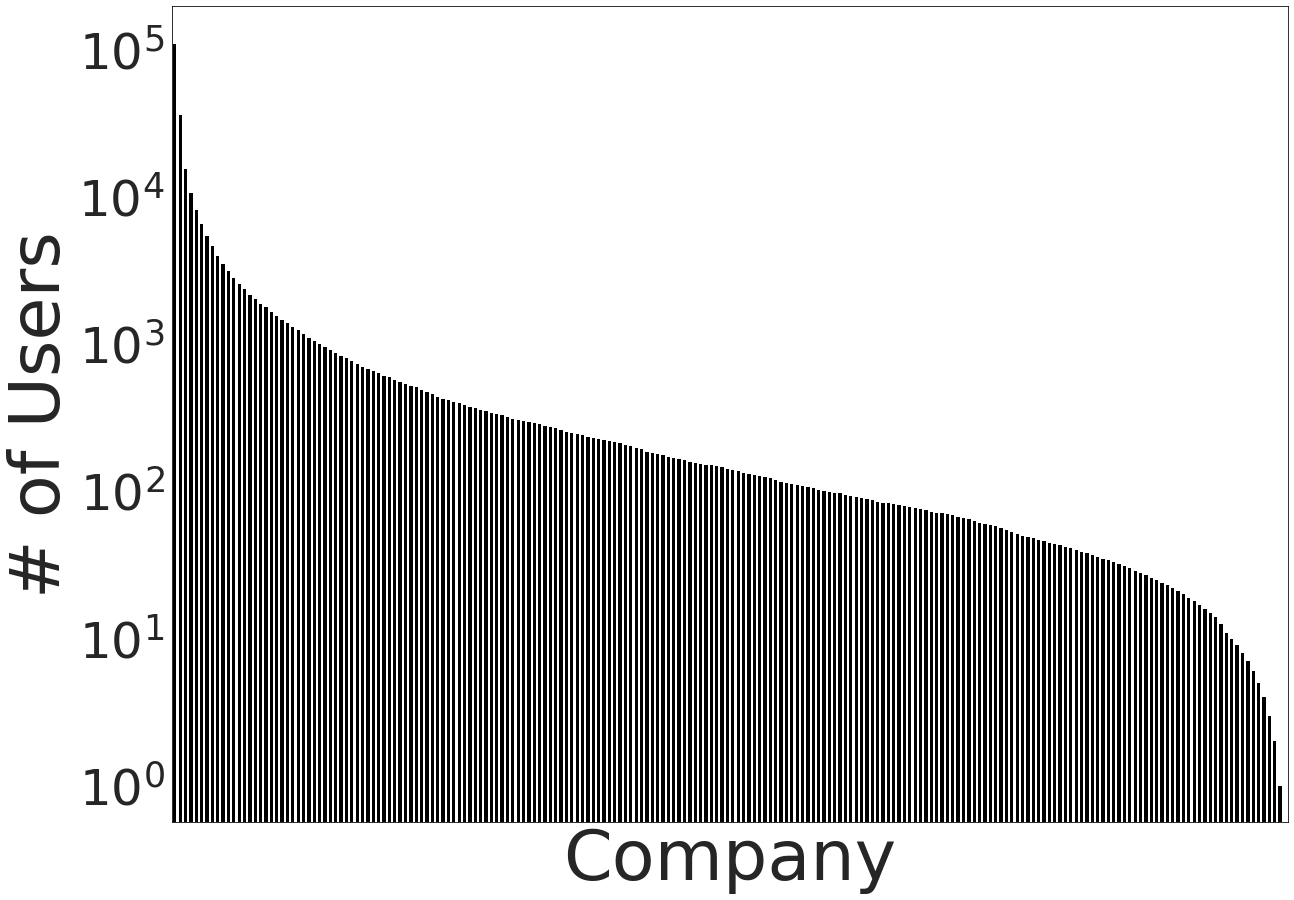}}
    \caption{Distributions of our dataset (\ie, $x$-axis indicates positions or companies in frequency order, while $y$-axis indicates the number of users in log scale).}
    \label{fig:data_distribution}
\end{figure}

After standardizing the position names, the distribution of the normalized names is shown in Figure \ref{fig:data_distribution}-(a).
On the other hand, Figure~\ref{fig:data_distribution}-(b) shows the distribution of company names.
In addition, Table \ref{tab:job_stats} shows the top-10 most-frequent normalized position names. 

\begin{table}[t]
    \small
    \centering
    \caption{Top ten frequent positions in the dataset
    }
    \resizebox{0.4\textwidth}{!} {    
        \begin{tabular}{l|c}
        \toprule
        \multicolumn{1}{c|}{Normalized Position}&Percentile (\%)\\\hline
        \hline
        policy administration professionals&2.49\\\hline
        systems analysts&2.39\\\hline
        social work and counselling professionals&2.30\\\hline
        advertising and marketing professionals&2.21\\\hline
        personnel and careers professionals&2.07\\\hline
        university and higher education teachers&1.97\\\hline
        sales and marketing managers&1.89\\\hline
        retail and wholesale trade managers&1.80\\\hline
        office supervisors&1.73\\\hline
        administrative and executive secretaries&1.56\\
        \bottomrule
        \end{tabular}
    }
        \label{tab:job_stats}
\end{table}

Lastly, we conducted a detailed analysis by computing the average working duration between jobs for specific sectors. We filtered our dataset based on job titles, categorizing them into distinct domains such as Tech, Sales, University, and Medical. The criterion for filtering included specific keywords within job titles, such as ``software'' for Tech and ``university'' for University. 
Below, we provide examples from four domains that we examined: 

\begin{itemize}
    \item Tech: Mean duration of 42.8 months, Std of 22.9 months
    \item Sales: Mean duration of 43.1 months, Std of 22.1 months
    \item University: Mean duration of 43.8 months, Std of 27.5 months
    \item Medical: Mean duration of 43.4 months, Std of 23.8 months
\end{itemize}

The results revealed that while the mobility in the Tech sector is marginally higher, the differences across these domains are not statistically significant ($p<0.01$). This suggests a relatively uniform mobility pattern across different professions. Also, it is important to note that career advancement and promotion opportunities exist within the same company or organization, which is also common in university-related jobs.

\section{Implementation Details}\label{appendix:parameters}
For hyperparameters of competitors, we used the best settings found via grid search in the following ranges: embedding dimensionality $\in\{50, 100, 150, 200, 250, 300\}$ (for DaeMon, CENET, NEMO, HCPNN, AHEAD, TACTP, and NAOMI) and the number of GCN layers $\in\{1,2,3\}$ (for AHEAD). 
Therefore, the optimal dimensionalities for embedding were 100 for DaeMon, NEMO, HCPNN, and AHEAD, 150 for TACTP, 200 for CENET, and 250 for NAOMI.
Also, based on our experiments, the most optimal number of GCN layers for AHEAD is 1.

For \ours, we set the embedding dimensionality, the number of GCN layers, and the max epoch for training as 150, 1, and 100, respectively. 
When we employ the LSTMs in (M3), \texttt{GRNN} can be formulated to obtain $\tilde{\mathbf{u}}_x(t_m)$ of $u_x$ with respect to $t_m$:
\begin{equation}\label{eq:lstm}
\begin{aligned}
&\mathbf{f}_{u_x}(t_m) = \mathbf{W}_{f_{(u,1)}}(\tilde{\mathbf{u}}_x(t_{m-1})) + \mathbf{W}_{f_{(u,2)}}(\mathbf{u}_x(t_m)),\\
&\mathbf{i}_{u_x}(t_m) = \mathbf{W}_{i_{(u,1)}}(\tilde{\mathbf{u}}_x(t_{m-1})) + \mathbf{W}_{i_{(u,2)}}(\mathbf{u}_x(t_m)),\\
&\mathbf{g}_{u_x}(t_m) = \mathbf{W}_{g_{(u,1)}}(\tilde{\mathbf{u}}_x(t_{m-1})) + \mathbf{W}_{g_{(u,2)}}(\mathbf{u}_x(t_m)),\\
&\mathbf{o}_{u_x}(t_m) = \mathbf{W}_{o_{(u,1)}}(\tilde{\mathbf{u}}_x(t_{m-1})) + \mathbf{W}_{o_{(u,2)}}(\mathbf{u}_x(t_m)),\\
&\mathbf{cell}_{u_x}(t_m) = \mathbf{cell}_{u_x}(t_{m-1}) \circ \mathbf{f}_{u_x}(t_m) + \mathbf{i}_{u_x}(t_m) \circ \mathbf{g}_{u_x}(t_m),\\
&\tilde{\mathbf{u}}_x(t_m) = \mathbf{o}_{u_x}(t_m) \circ \tanh(\mathbf{cell}_{u_x}(t_m)),\\
\end{aligned}
\end{equation}
where $\mathbf{cell}_{u_x}(t_m)$ indicates a cell state for $\tilde{\mathbf{u}}_x(t_m)$ of $u_x$ with respect to $t_m$. 
Also, $\mathbf{f}_{u_x}(t_m), \mathbf{i}_{u_x}(t_m), \mathbf{g}_{u_x}(t_m),$ and $\mathbf{o}_{u_x}(t_m)$ represent the intermediate calculation processes to obtain $\tilde{\mathbf{u}}_x(t_m)$.
Lastly, $\mathbf{W}_{f_{(u,\cdot)}}, \mathbf{W}_{i_{(u,\cdot)}}, \mathbf{W}_{g_{(u,\cdot)}},$ and  $\mathbf{W}_{o_{(u,\cdot)}}$ represent the learnable weight matrices; they are shared across all users.
In the same manner, we can obtain $\tilde{\mathbf{c}}_z(t_m)$ of $c_z$ with respect to $t_m$.

\section{Data Sharing Statement}\label{appendix:sharing}

The rarity of accessible job/resume datasets underscores the significance of our commitment to reproducibility. 
It is noteworthy that none of the over 30 papers~\cite{abs-1609-06268, ZhangXZQMXC19, 
zhang2019job2vec, YamashitaTL24, yamashita2023james, XuLGAB14, XuYYXZ18, WangAKB19, WangZPB13, TengZZLX19, SunZXSX19, ShiYGH20, ShiSYKCH20, RamanathIPGKO18, QinHRSKZL20, QinZXCZC18, PaparrizosCG11, OentaryoLAPOL18, MaheshwariSR10, luo2019learning, LiuZNYR16, LiYZTL23, LiSYWCH20, LiZPZ16, LiGZXZ17, DaveZHKK18, DaiYQX0C020,li2017nemo,meng2019hierarchical,zhang2021attentive,wang2021variable,yamashita2022looking} that propose algorithms using job/resume datasets have opted to share their datasets due to the sensitive nature of these datasets. 
Privacy concerns surrounding resume datasets, which contain sensitive personal information, further hinder their public release. Consequently, the majority of CTP literature, including studies such as NEMO \cite{li2017nemo}, HCPNN \cite{meng2019hierarchical}, AHEAD \cite{zhang2021attentive}, TACTP \cite{wang2021variable}, and NAOMI \cite{yamashita2022looking}, has relied on a single private dataset, none of which have been publicly shared. The proprietary nature of the datasets in the CTP domain makes finding publicly available datasets very challenging. 

As a result, we had to rely solely on our own dataset, obtained from a global career platform, for evaluation. 
However, with explicit consent from the company that donated our dataset, we pledge to make it available upon request, with all due legal protocols observed. 
Our dedication to openness extends beyond mere rhetoric—we have also shared our entire codebase. 
For the benefit of the research community, 
our dataset is accessible upon request; interested researchers should contact FutureFit AI or the Penn State team to access the original dataset.
Additionally, in~\cite{Yamashita24}, we introduced OpenResume, an anonymized, structured, and publicly available resume dataset, specifically curated for downstream tasks in the job domain.

\end{document}